\documentclass[10pt,twocolumn,letterpaper]{article}

\usepackage{iccv}
\usepackage{times}
\usepackage{epsfig}
\usepackage{graphicx}
\usepackage{amsmath}
\usepackage{amssymb}

\usepackage{graphicx}
\usepackage{amsmath}
\usepackage{amssymb}
\usepackage{booktabs}

\usepackage[table]{xcolor}
\definecolor{bestcolor}{rgb}{1.0,1.0,1.0} 
\definecolor{secondcolor}{rgb}{1.0,1.0,1.0} 
\definecolor{thirdcolor}{rgb}{1.0,1.0,1.0} 

\usepackage[pagebackref=true,breaklinks=true,letterpaper=true,colorlinks,bookmarks=false]{hyperref}

\usepackage{url}            
\usepackage{booktabs}       
\usepackage{amsfonts}       
\usepackage{times}
\usepackage{epsfig}
\usepackage{amsmath}
\usepackage{amssymb}
\usepackage{multirow}
\usepackage{subcaption}
\usepackage{cuted}
\usepackage{changepage}
\usepackage{amsthm}
\usepackage{xcolor}
\usepackage{enumitem}
\usepackage{tikz}

\usepackage[capitalize]{cleveref}
\crefname{section}{Sec.}{Secs.}
\Crefname{section}{Section}{Sections}
\Crefname{table}{Table}{Tables}
\crefname{table}{Tab.}{Tabs.}
\usepackage{appendix}

\newcommand{\methodname}{Strivec}

\iccvfinalcopy 

\def\iccvPaperID{3306} 

\ificcvfinal\fi

\newcommand{\boldstartspace}[1]{\vspace{0.05in}\noindent\textbf{#1}}

\newcommand{\first}{\tikz\draw[yellow,fill=yellow] (0,0) circle (0.7ex);}
\newcommand{\second}{\tikz\draw[lightgray,fill=lightgray] (0,0) circle (0.7ex);}
\newcommand{\third}{\tikz\draw[brown,fill=brown] (0,0) circle (0.7ex);}

\begin{document}

\title{Strivec: Sparse Tri-Vector Radiance Fields}

\author{Quankai Gao$^{*1}$ \qquad Qiangeng Xu$^{*1}$ \qquad Hao Su$^{2}$ \qquad Ulrich Neumann$^{1}$ \qquad Zexiang Xu$^{3}$  \\
    \hspace{0mm}$^1$University of Southern California \hspace{18mm} 
    $^2$UC San Diego\hspace{18mm}
    $^3$Adobe Research\\
    {\tt\small \hspace{0mm}\{quankaig,qiangenx,uneumann\}@usc.edu}\hspace{5mm}
    {\tt\small haosu@ucsd.edu}\hspace{5mm}
    {\tt\small zexu@adobe.com}\qquad
}

\maketitle
\ificcvfinal\thispagestyle{empty}\fi
\maketitle
\let\thefootnote\relax\footnotetext{$^*$Equal contribution.\\
Code and results: \href{https://github.com/Zerg-Overmind/Strivec}{https://github.com/Zerg-Overmind/Strivec}}

\begin{abstract}
   We propose \methodname{}, a novel neural representation that models a 3D scene as a radiance field with sparsely distributed and compactly factorized local tensor feature grids.
   Our approach leverages tensor decomposition, following the recent work TensoRF~\cite{chen2022tensorf}, to model the tensor grids.  In contrast to TensoRF which uses a global tensor and focuses on their vector-matrix decomposition,  we propose to utilize a cloud of local tensors and apply the classic CANDECOMP/PARAFAC (CP) decomposition \cite{carroll1970analysis} to factorize each tensor into triple vectors that express local feature distributions along spatial axes and compactly encode a local neural field.
   We also apply multi-scale tensor grids to discover the geometry and appearance commonalities and exploit spatial coherence with the tri-vector factorization at multiple local scales.  The final radiance field properties are regressed by aggregating neural features from multiple local tensors across all scales. Our tri-vector tensors are sparsely distributed around the actual scene surface, discovered by a fast coarse reconstruction, leveraging the sparsity of a 3D scene.
   We demonstrate that our model can achieve better rendering quality while using significantly fewer parameters than previous methods, including TensoRF and Instant-NGP~\cite{muller2022instant}.

\end{abstract}


\begin{figure}[t]
        \centering
        \setlength{\abovedisplayskip}{0pt}%
        \setlength{\abovedisplayshortskip}{\abovedisplayskip}%
        \setlength{\belowdisplayskip}{0pt}%
            \begin{center}{}
                \centering
                \includegraphics[width=1.\linewidth]{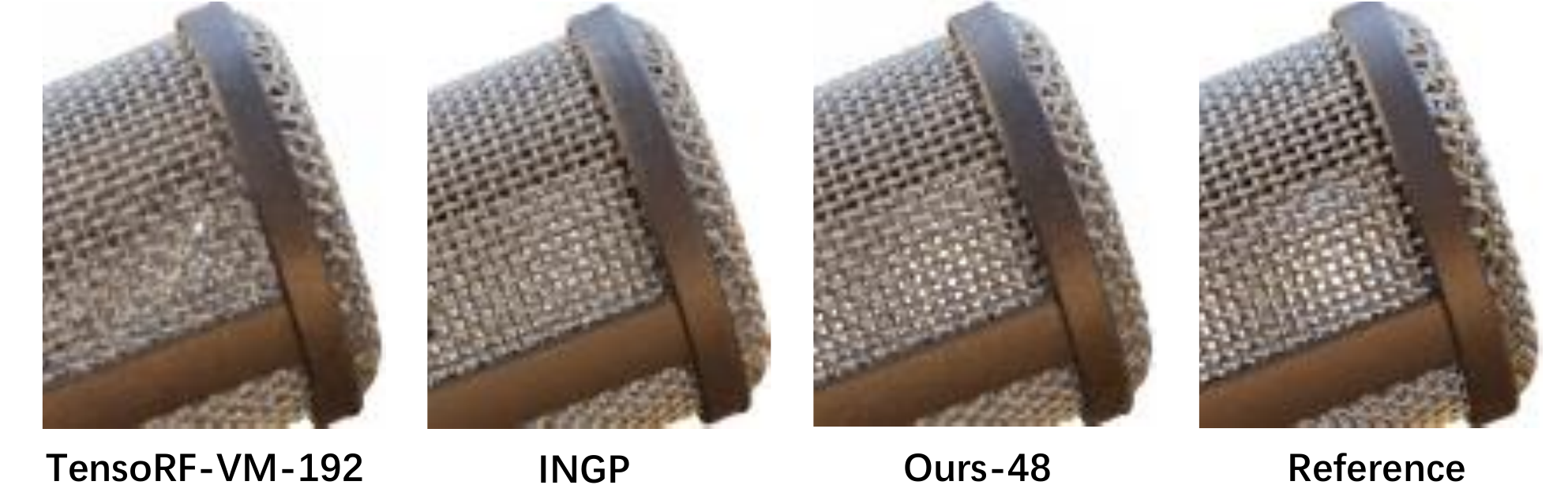}
                
                
                \includegraphics[width=1\linewidth]{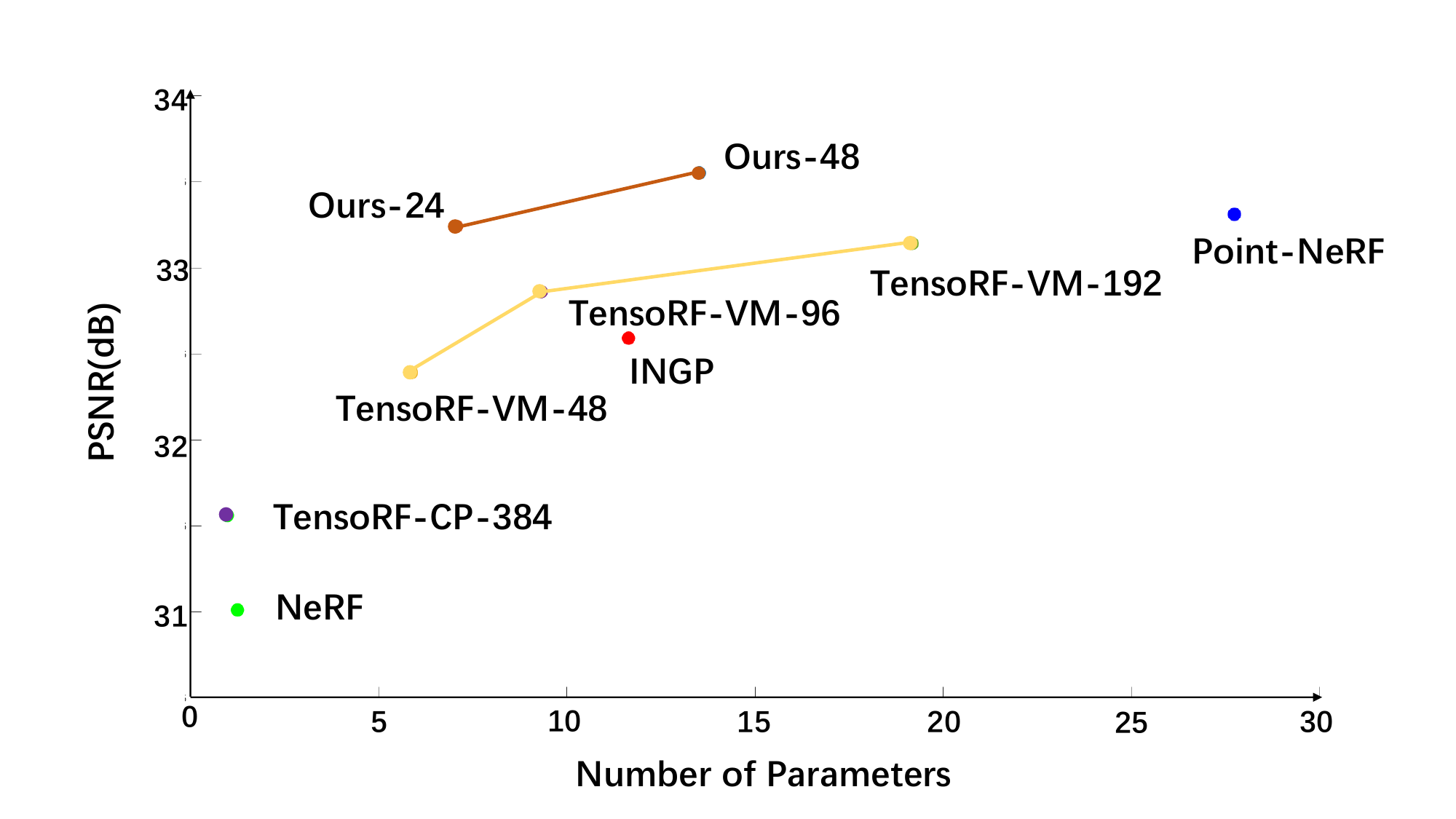} 
                
            \end{center} 
        \captionsetup{aboveskip=3pt}
        \captionsetup{belowskip=-5pt}
        \vspace{-0pt}
        \caption{
        We compare with previous methods in terms of rendering quality (PSNR) and model capacity (number of parameters) on the NeRF Synthetic dataset on the bottom. 
        Our method and TensoRF are shown with different model sizes. 
        Our approach consistently achieve better rendering quality with fewer model parameters than TensoRF, as well as other methods like iNGP. On the top, we show one example of visual comparisons of the mic scene that has challenging fine-grained geometric structures, where our approach captures most of the details and is the closest to the reference. Note that the results of NeRF and Point-NeRF use 200k optimization steps while the rest use only 30k steps.}
        \label{fig:teaser}
    \end{figure}

\section{Introduction}

Representing 3D scenes as radiance fields \cite{mildenhall2020nerf} has enabled photo-realistic rendering quality and emerged as a popular design choice in 3D vision and graphics applications.  
While many methods \cite{park2021nerfies,zhang2020nerf++,barron2021mip} (following NeRF \cite{mildenhall2020nerf}) purely use MLPs to represent neural fields, 
recent works, like TensoRF \cite{chen2022tensorf} and Instant-NGP \cite{muller2022instant}, have demonstrated the advantages of using shared global feature encoding for radiance field modeling, in terms of speed, compactness, and quality.
However, these methods share and assign neural features uniformly in a scene (with tensor factors or hash tables), assuming the scene content is equally complex over the entire space, which can be inefficient (requiring high model capacity) to accurately model intricate local scene details (see Fig.\ref{fig:teaser}).

We aim to accurately and compactly model a 3D scene and reproduce the complex local details.
To this end, we propose \methodname{}, a novel neural scene representation that utilizes \emph{sparsely distributed} and \emph{compactly factorized} local tensor grids to model a volumetric radiance field for high-quality novel view synthesis. 
As shown in Fig.\ref{fig:teaser}, our approach is able to accurately model the complex scene structures that are not recovered well by previous methods. 
More importantly, our superior rendering quality is achieved with much less model capacity.

In particular, we base our model on TensoRF \cite{chen2022tensorf}, a recent approach that leverages tensor factorization in radiance field modeling. It is fast, compact, and of high rendering quality. 
TensoRF applies CP and vector-matrix (VM) decomposition techniques to factorize a field into vectors and matrices and model the entire scene as a global factorized tensor.
Instead of a single global tensor, we leverage a sparse set of multiple small local tensors distributed around the scene surface for more efficient scene modeling.
Specifically, each of our tensors represents a local radiance field inside its local bounding box and is compactly modeled with factorized triple vectors based on the CP decomposition.

Note that the global CP decomposition in TensoRF has led to a highly compact model but cannot achieve comparable rendering quality to their VM decomposition.
This is because a tri-vector CP component is rank-one, while a global feature grid of an entire 3D scene is often complex and of high rank, requiring a large (impractical) number of CP components for high accuracy.
TensoRF addresses this by introducing matrix factors in their VM decomposition, essentially increasing the rank of each tensor component.
Our model instead consists of multiple small tensor grids, exploiting local spatial commonalities in a scene. 
Compared to a global tensor, our local tensor is less complex and of much lower rank, thus effectively reducing the required number of CP components (per tensor) and enabling practical high-quality radiance field reconstruction with
 highly compact tri-vector factors.
 Our local tri-vector tensors can lead to superior rendering quality and compactness over TensoRF's VM model (see Fig.~\ref{fig:teaser}). We also observe that our local tensors are generally more robust than a global tensor against the orientation of spatial axes (which can affect the rank of a tensor and thus affects the quality; see Fig.~\ref{tb:rot}).

Importantly, adopting local tensors (instead of a global one)
also brings us the flexibility to allocate neural features according to the actual scene distribution, enabling more efficient scene modeling and better usage of model parameters than a global representation. 
To do so, we pre-acquire coarse scene geometry -- that can be easily achieved via a fast RGB$\sigma$ volume reconstruction (like DVGO \cite{sun2022direct}) or multi-view stereo (like Point-NeRF \cite{xu2022point}) -- to directly distribute local tensors around the actual scene surface, leading to a sparse scene representation that avoids unnecessarily modeling the empty scene space.
Note that while previous methods have also leveraged sparse representations (with voxels \cite{liu2020neural,yu2021plenoxels} or points \cite{xu2022point}) of radiance fields, their local features are modeled and optimized independently.  
Our model instead correlates a group of local features inside a local box and compactly express them with triple vectors, uniquely exploiting the local spatial coherence along axes and imposing local low-rank priors in the feature encoding via tensor factorization.
Moreover, unlike previous sparse representations that only use a single-scale feature grid or point cloud, we distribute multi-scale local tensors to effectively model the scene geometry and appearance at multiple scales in a hierarchical manner.
In particular, for an arbitrary 3D location, we aggregate the neural features from its neighboring tri-vector components at all scales and decode the volume density and view-dependent color from the aggregated features for radiance field rendering.

Our approach takes the best of previous local and global radiance field representations.
Compared with global representations like TensoRF and Instant-NGP, our model takes advantage of the sparsity of a scene more directly; compared with local representations like Plenoxels and Point-NeRF, our model makes use of the local smoothness and coherence of scene geometry and appearance. 
As shown in our experimental results on both synthetic and real datasets, our model is able to achieve state-of-the-art rendering quality on these datasets, outperforming previous methods, including TensoRF and Instant-NGP, while using significantly fewer model parameters, demonstrating the superior representational power of our model.

\section{Related Work}
\paragraph{Scene representations.}
To represent a 3D scene, traditional and learning-based methods have studied various representations, such as depth map \cite{huang2018deepmvs,liu2015learning}, mesh \cite{kanazawa2018learning,wang2018pixel2mesh,shewchuk1998tetrahedral}, point cloud \cite{qi2017pointnet,achlioptas2018learning,wang2018mvpnet} and implicit function \cite{chen2018learning,mescheder2018occupancy,niemeyer2020differentiable,yariv2020multiview}. 
In recent years, continuous neural field representations stand out in various 3D tasks such as single-view 3D reconstruction \cite{xu2019disn,genova2020local}, surface completion \cite{chibane2020implicit, park2019deepsdf}, multi-view reconstruction \cite{niemeyer2020differentiable} and novel view synthesis \cite{mildenhall2020nerf,martin2021nerf}. 
Compared with traditional discrete representations, a continuous field have no limitation on spatial resolution, e.g., volume resolution or the number of points. 
It can also naturally be represented by neural networks, such as an MLP, which are known for approximating complex functions well. 

\paragraph{Neural field representations.}
Specifically, NeRF \cite{mildenhall2020nerf} represents a 3D scene as a radiance field with a global coordinate MLP, which models geometry, lighting and texture information jointly, leading to photo-realistic rendering quality in novel view synthesis. 
Apart from its advantage, purely MLP-based NeRF models \cite{barron2021mip,verbin2022ref} in general suffer from inefficiency \cite{arandjelovic2021nerf} when modeling highly complex or large-scale scenes, due to limited model capacity, slow optimization speed, and the cost of modeling empty space.

To model radiance fields more efficiently, recent works have explored combining neural fields with various traditional 3D representations, including voxels \cite{liu2020neural,yu2021plenoxels,sun2022direct,zhang2022nerfusion} and points \cite{xu2022point}. Low-rank representations such as triplane~\cite{chan2022efficient,fridovich2023k} and tensor decomposition~\cite{chen2022tensorf,obukhov2022tt} have also been studied.
In particular, DVGO \cite{sun2022direct} and Plenoxels \cite{yu2021plenoxels} respectively use dense and sparse voxels with neural features for radiance field modeling.
While being efficient to optimize, these localized feature grid-based representations lead to a large model size and can face overfitting issues when the features are of very high resolution.
Consequently, DVGO can also work with a low-resolution grid and Plenoxels requires additional spatial regularization terms.
On the other hand, recent works have adopted global feature encoding to express a high-resolution feature grid, including Instant-NGP \cite{muller2022instant} that hashes spatial features into multi-scale hash tables and TensoRF \cite{chen2022tensorf} that factorizes a feature grid into vector and matrix factors.
These global feature encoding methods exploit the spatial correlation across the entire scene space, leading to fast and compact reconstruction and surpassing previous MLP-based or grid-based representations on rendering quality.
However, similar to NeRF, such global representation can also be limited by its model capacity when representing highly complex or large-scale content.

Our approach instead combines local and global representations.
Our tri-vector fields are sparsely distributed in the scene, similar to local representations (like plenoxels and Point-NeRF); meanwhile, features in each field are represented by tri-vector components shared across the local region as done in TensoRF, exploiting spatial feature commonalities.
Our model leverages both spatial sparsity and coherence, leading to much higher compactness and better reconstruction quality than previous local and global representations (see Tab.~\ref{tb:nerfsynth}).

Relevant to our work, previous methods, such as KiloNeRF \cite{reiser2021kilonerf} and BlockNeRF \cite{tancik2022block} have also utilized multiple local MLPs to represent a scene.
Specifically, KiloNeRF focuses and speeding up NeRF and their rendering quality is sacrificed; BlockNeRF essentially uses multiple NeRFs to increase the total model capacity.
Instead of pure MLPs, our work is built upon tensor factorization-based feature encoding as done in TensoRF \cite{chen2022tensorf}, and we in fact achieve superior rendering quality while decreasing the model capacity.

\newcommand{\Img}{I}
\newcommand{\Cam}{\Phi}
\newcommand{\iI}{q}

\newcommand{\Color}{c}
\newcommand{\ShadX}{x}
\newcommand{\iS}{q}
\newcommand{\iSS}{t}
\newcommand{\ShadNum}{Q}

\newcommand{\Dir}{\mathbf{d}}
\newcommand{\Rad}{c}
\newcommand{\Trans}{T}
\newcommand{\Dens}{\sigma}
\newcommand{\Step}{\delta}
\newcommand{\Contr}{\alpha}

\newcommand{\TensorThreeD}{\tau}
\newcommand{\TensorCloud}{\mathcal{T}}

\newcommand{\TensorNum}{N}
\newcommand{\TensorX}{p}
\newcommand{\TensorL}{l}

\newcommand{\iT}{n}
\newcommand{\iLT}{m}
\newcommand{\TensorF}{f}
\newcommand{\TensorV}{\mathbf{v}}
\newcommand{\TensorWeight}{w}
\newcommand{\TensorSpace}{\omega}
\newcommand{\TensorGrid}{\mathcal{G}}
\newcommand{\TensorCompNumber}{R}
\newcommand{\TensorComp}{\mathcal{A}}

\newcommand{\CoveredSpace}{\Omega}
\newcommand{\ShadingX}{\chi}
\newcommand{\QueryNum}{M}

\newcommand{\ScaleNum}{S}
\newcommand{\iScale}{s}

\newcommand{\Densf}{f^\Dens}
\newcommand{\Radf}{f^\Rad}
\newcommand{\AggF}{A}
\newcommand{\AppearanceM}{\mathbf{B}}
\newcommand{\AppearanceMC}{\AppearanceM^\Rad}
\newcommand{\AppearanceV}{\mathbf{b}}
\newcommand{\DensShift}{\epsilon}

\newcommand{\AppearanceFunc}{\psi}
\newcommand{\GTColor}{\Tilde{C}}

\newcommand{\ThresContr}{T_{\text{opacity}}}
\newcommand{\ThresDist}{T_{\text{dist}}}

\begin{figure*}[!hbt]
        \centering
        \setlength{\abovedisplayskip}{0pt}%
        \setlength{\abovedisplayshortskip}{\abovedisplayskip}%
        \setlength{\belowdisplayskip}{0pt}%
            \begin{center}{}
                \centering
                \includegraphics[width=1\linewidth]{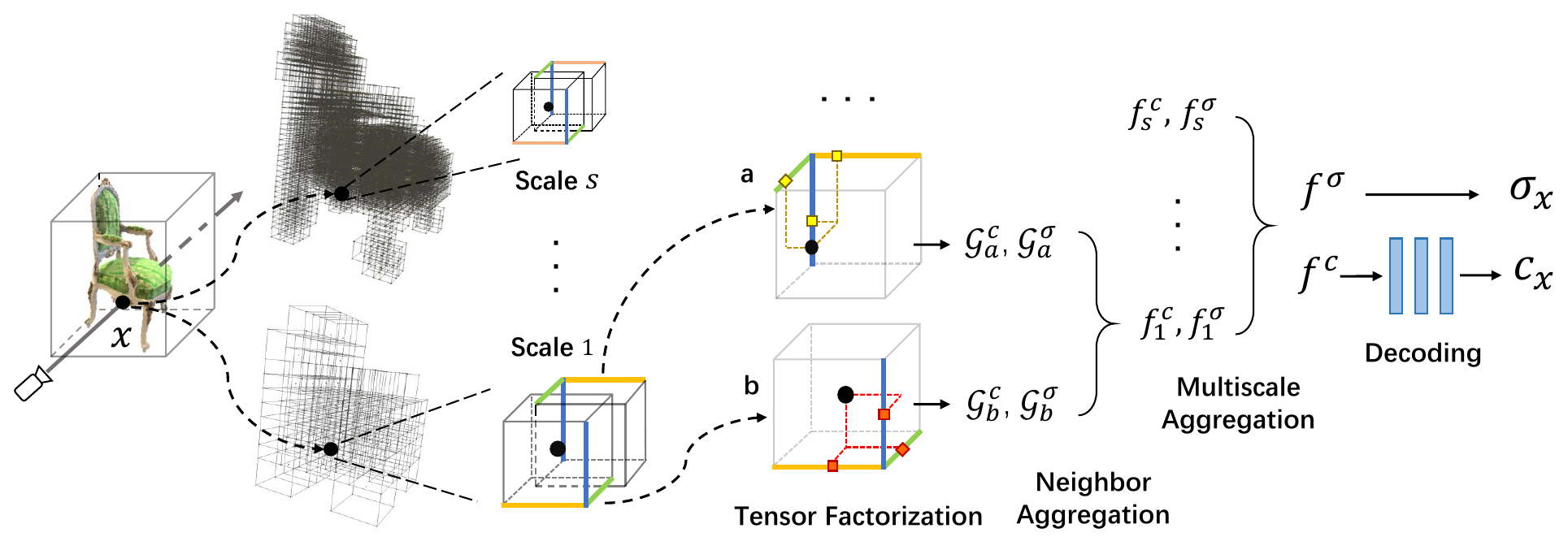} 
            \end{center} 
        \captionsetup{aboveskip=0pt}
        \captionsetup{belowskip=-5pt}
        \caption{Overview of our Sparse Tri-Vector Radiance Fields. We distribute our local tensors based on a coarse geometry estimated by a fast RGB$\sigma$ volume reconstruction as done in DVGO~\cite{sun2022direct}. Here, we show our model running under $\ScaleNum=2$ scales. Each local tensor is factorized as axis-aligned triple based on CP decomposition. For any shading point $\ShadingX$, we extract and evaluate features in each local tensor, according to the factorization (Sec.~\ref{sec:localtri}). Then, we aggregate these features among nearby tensors (Sec.~\ref{sec:feaAgg}) and across different scales (Sec.~\ref{sec:scaleAgg}). Finally, the density and color are decoded (Sec.~\ref{sec:decoding}) and used by volume rendering (Sec.\ref{sec:renrecon}).} 
        \label{fig:overview}
    \end{figure*}
    
    
\section{Sparse Tri-Vector Field Representation}
We now present our novel radiance field representation.
In essence, our model consists of a cloud of small local tri-vector tensors at multiple scales, designed to leverage both sparsity and multi-scale spatial coherence (see Fig. \ref{fig:overview}).

Let $\TensorCloud = \{\TensorThreeD_\iT|\iT=1,...,\TensorNum\}$ denote a cloud of tri-vector tensors. 
Each local tensor $\TensorThreeD$ is located at $\TensorX$, covering a local cuboid space $\TensorSpace$ with an edge length of $\TensorL$. 
This cloud of tri-vector tensors represents a radiance field for the 3D space:
    \begin{align}
        \CoveredSpace = \bigcup_{\iT=1}^{N}\TensorSpace_{\iT}.
    \end{align}

Here, each tensor $\TensorThreeD$ encodes a local multi-channel feature grid that includes a (single-channel) density grid $\AggF_\Dens$ and a (multi-channel) appearance grid $\AggF_\Rad$, similar to the tensor grid in TensoRF \cite{chen2022tensorf}.
In contrast to using a single global tensor in TensoRF \cite{chen2022tensorf}, we model the volume density and view-dependent colors with multiple local tensors.
In particular, for an arbitrary location $\ShadingX \in \CoveredSpace$, we select $\QueryNum$ nearest tensors that cover $\ShadingX$. 
Across the selected tensors, we aggregate the extracted density and appearance features recovered by their tri-vector factors for radiance field property regression, where the volume density $\Dens$ is directly obtained after the aggregation and the view-dependent color $\Rad$ is regressed by a small MLP $\AppearanceFunc$ along with the viewing direction $\Dir$.
The continuous radiance field can be expressed as: 
    \begin{align}
        \Dens_\ShadingX, \Rad_\ShadingX = \AggF_\Dens(\{\TensorGrid^\Dens(\ShadingX)\}), \AppearanceFunc(\AggF_\Rad(\{\TensorGrid^\Rad(\ShadingX)\}), \Dir).
    \end{align}

\subsection{Local tri-vector tensors.}
\label{sec:localtri}
We apply the classic Canonical polyadic (CP) decomposition~\cite{carroll1970analysis} to model our local tensors with tri-vector components.

\paragraph{CP decomposition.} 
CP decomposition factorizes a $M$ dimension tensor $\mathcal{\TensorThreeD} \in \mathbb{R}^{I_1 \times I_2 \times ... \times I_M}$ into a linear combination of $\TensorCompNumber$ rank-1 tensors:
    \begin{equation}
    \begin{aligned}
        \mathcal{\TensorThreeD} &= \sum^\TensorCompNumber_{r=1}{\lambda_r \TensorV^0_r \otimes \TensorV^1_r \otimes ..., \otimes \TensorV^M_r},
    \label{eq:cp}
    \end{aligned}
    \end{equation}
where $\otimes$ denotes outer product; the weighting factor $\lambda_r$ can be absorbed into vectors $\{\TensorV_r^0, ..., \TensorV_r^M\}$.

\paragraph{Density and appearance tensors.}
In our case of modeling a 3D radiance field, we set the geometry grid $\TensorGrid^\Dens \in \mathbb{R}^{I \times J \times K}$ as a 3D tensor. And the multi-channel appearance grid $\TensorGrid^\Rad \in \mathbb{R}^{I \times J \times K \times P}$ corresponds to a 4D tensor. The fourth appearance mode is of lower dimension (compared with the spatial modes), representing the final dimension of the features sent to the MLP decoder network.

According to Eqn.\ref{eq:cp}, we factorize each tensor's feature grids, $\TensorGrid^\Dens$ and $\TensorGrid^\Rad$, by CP decomposition:
    \begin{align}
        \TensorGrid^\Dens &= \sum^{\TensorCompNumber_\Dens}_{r=1}{\TensorComp_{\Dens,r}} = \sum^{\TensorCompNumber_\Dens}_{r=1}{\TensorV^X_{\Dens,r} \otimes \TensorV^Y_{\Dens,r} \otimes \TensorV^Z_{\Dens,r}},
        \label{eq:denfact} \\
        \TensorGrid^\Rad &= \sum^{\TensorCompNumber_\Rad}_{r=1}{\TensorComp_{\Rad,r} \otimes \AppearanceV_r} = \sum^{\TensorCompNumber_\Rad}_{r=1}{\TensorV^X_{\Rad,r} \otimes \TensorV^Y_{\Rad,r} \otimes \TensorV^Z_{\Rad,r} \otimes \AppearanceV_r},
         \label{eq:appfact} 
    \end{align}
Here $R_\Dens$ and $R_\Rad$ denote numbers of component; $\TensorComp_{\Dens,r}$ and $\TensorComp_{\Rad,r}$ are the component tensors that are factorized spatially; $\TensorV^X_{\Dens,r}, ..., \TensorV^X_{\Rad,r}, ...$ are the 1D vectors with resolution $I,J,K$, modeling scene geometry and appearance along $X,Y,Z$ axis; $\TensorCompNumber_\Dens$ and $\TensorCompNumber_\Rad$ are the component numbers;  $\AppearanceV_r$ expresses the feature dimension.

As done in TensoRF \cite{chen2022tensorf}, we stack all feature-mode vectors $\AppearanceV_r$ as columns together, which ends up a $P \times \TensorCompNumber_\Rad$ appearance matrix $\AppearanceM$. 
This matrix models the appearance feature variations of the tensor and functions like a appearance dictionary. 
Note that naively following CP decomposition like TensoRF will assign a different appearance matrix for every local tensor.
Instead, we propose to utilize a global appearance matrix $\AppearanceMC$ shared across the entire cloud of local tensors, leading to a global appearance dictionary that explains the color correlations across scene. This further improves both the computational efficiency and model compactness of our model.

Therefore, each of our local tensors is represented by their unique local tri-vector factors $\TensorV^X_{r}$, $\TensorV^Y_{r}$, $\TensorV^Z_{r}$.

\paragraph{Feature evaluation.}
To achieve a continuous field, we consider trilinear interpolation when evaluating the tensor grid features. 
For a location $\ShadingX$, we first compute its relative position $\Tilde{\ShadingX}$ to the selected tensor located at $\TensorX$:
    \begin{align}
        \Tilde{x},\Tilde{y},\Tilde{z} = x - \TensorX_x, y - \TensorX_y, z - \TensorX_z.
        \label{eq:relapos} 
    \end{align}
Then, for example, to get $\TensorComp_{\Dens,r}$ at ($\Tilde{x},\Tilde{y},\Tilde{z}$), we can compute and trilinearly interpolate eight $\TensorComp_{\Dens,r}$ on the corners. As mentioned in \cite{chen2022tensorf}, applying linear interpolation on each vector first is mathematically equivalent and can reduce the computation cost. 
Under the rule of outer product, we have $\TensorComp_{r,i,j,k} = \TensorV_{r,i}^X \TensorV_{r,j}^Y \TensorV_{r,k}^Z$, then the interpolated density features at location $\ShadingX$ are:
    \begin{align}
         \TensorGrid^\Dens(\ShadingX) & = \sum_r{\TensorV_{\Dens,r}^X(\Tilde{x}) \TensorV_{\Dens,r}^Y(\Tilde{y}) \TensorV_{\Dens,r}^Z(\Tilde{z})} = \sum_r{\TensorComp_{\Dens,r}(\Tilde{\ShadingX})},
         \label{eq:interp} 
    \end{align}
where $\TensorV_{\Dens,r}^X(\Tilde{x})$ is $\TensorV_{\Dens,r}^X$'s linearly interpolated value at $(\Tilde{x})$ along its $X$ axis. Here, $\TensorGrid^\Dens(\ShadingX)$ is a scalar.

Similarly, the interpolated appearance features can be computed as:
    \begin{align}
         \TensorGrid^\Rad(\ShadingX) &= \sum_r{\TensorV_{\Rad,r}^X(\Tilde{x}) \TensorV_{\Rad,r}^Y(\Tilde{y}) \TensorV_{\Rad,r}^Z(\Tilde{z}) \AppearanceV_r} \\
                                      &= \sum_r{\TensorComp_{\Rad,r}(\Tilde{\ShadingX})\AppearanceV_r} \\
                                      &= \AppearanceM \cdot (\oplus [\TensorComp_{\Rad,r}]_r), 
         \label{eq:interp}
    \end{align}
where ``$\oplus$'' denotes concatenation, ``$\cdot$'' denotes dot product. The appearance feature $\TensorGrid^\Rad(\ShadingX) \in \mathbb{R}^P $ is a vector.

\subsection{Feature aggregation.}
\label{sec:feaAgg}
We propose to aggregate the features from $\QueryNum$ neighboring tensors to jointly model the volume density and appearance for each 3D location $\ShadingX$. 
In particular, inspired by Point-NeRF, we leverage an inverse distance-based weighting function to directly aggregate the multi-tensor features.
Specifically, this weight can be expressed by
    \begin{align}
        \TensorWeight_\iLT=\frac{1}{\lVert \TensorX_\iLT-\ShadingX \rVert}.
    \end{align}

With this weight function, we directly obtain the density feature via the weighted sum:
    \begin{align}
        f^\Dens(\ShadingX) = \frac{1}{\sum \TensorWeight_\iLT} \sum_{\iLT=1}^\QueryNum \TensorWeight_\iLT \TensorGrid_\iLT^\Dens(\ShadingX).
    \end{align}
Similarly, the appearance feature aggregation can be expressed in a similar way, while using the shared appearance matrix (as described in Sec.~\ref{sec:localtri}) across local tensors:
    \begin{align}
        f^\Rad(\ShadingX) &= \frac{1}{\sum \TensorWeight_\iLT} \sum_{\iLT=1}^\QueryNum \TensorWeight_\iLT \TensorGrid_\iLT^\Rad(\ShadingX) \\
               &= \frac{1}{\sum \TensorWeight_\iLT} \sum_{\iLT=1}^\QueryNum \TensorWeight_\iLT \AppearanceMC \cdot (\oplus [\TensorComp_{\Rad,r}]_r) \label{eq:btsorf} \\
               &= \frac{1}{\sum \TensorWeight_\iLT}\AppearanceMC \cdot (\sum_{\iLT=1}^\QueryNum \TensorWeight_\iLT (\oplus [\TensorComp_{\Rad,r}]_r)) \label{eq:bours}.
    \end{align}
Note that owing to sharing the appearance matrix across tensors, we reduce the computational complexity from $O(\QueryNum \cdot P \cdot \TensorCompNumber_\Rad)$ in Eqn.\ref{eq:btsorf}, to $O( (\QueryNum + P) \cdot \TensorCompNumber_\Rad )$ in Eqn.\ref{eq:bours}.

\subsection{Multi-scale tri-vector fields.}
\label{sec:scaleAgg}
Complex 3D scenes often contain multi-frequency geometry and appearance details. 
This motivates us to build multi-scale tensor clouds to discover the local geometry and appearance commonalities at multiple scales.
Our final radiance field is modeled by multiple tri-vector tensor clouds at $\ScaleNum$ different scales. Different clouds consist of tensors with different resolutions. 

To regress the density and appearance at a location $\ShadingX$, we gather the density and appearance features from a set of tensor clouds that cover $\ShadingX$, $\{\TensorCloud_\iScale | 1 \leq \iScale \leq \ScaleNum, \ShadingX \in \CoveredSpace_\iScale\}$. 
Please note that tensor clouds of certain scales might not cover the location, so that $ \lVert\{\TensorCloud_\iScale\}\rVert \leq \ScaleNum$. We simply compute the mean features across these scales:
    \begin{align}
        f^\Dens(\ShadingX) = \frac{1}{\lVert\{\TensorCloud_\iScale\}\rVert}\sum_sf^\Dens_\iScale(\ShadingX), \\
        f^\Rad(\ShadingX) = \frac{1}{\lVert\{\TensorCloud_\iScale\}\rVert}\sum_sf^\Rad_\iScale(\ShadingX).
    \end{align}

Note that $f^\Dens(\ShadingX)$ and $f^\Rad(\ShadingX)$ are the final density and appearance features we aggregate across multiple scales and multiple neighboring tensors.
\subsection{Decoding.}
\label{sec:decoding}
We apply softplus activation on the density feature $f^\Dens(\ShadingX)$ to obtain the final volume density and regress the view-dependent color by sending the appearance feature $f^\Rad(\ShadingX)$ and the viewing direction $\Dir$ to the MLP decoder $\AppearanceFunc$.

\begin{table*}[!hbt]
    \begin{adjustwidth}{0pt}{0pt}  
    \centering
        \small{
            \begin{tabular}{ccccccccc}
            \hline 
            Method & BatchSize & Steps & Time$\downarrow$ & \# Param.(M)$\downarrow$  & PSNR$\uparrow$ & SSIM$\uparrow$& LPIPS$_{Vgg}$ $\downarrow$ & LPIPS$_{Alex}$ $\downarrow$\\ \hline
            NeRF\cite{martin2021nerf} & 4096 & 300k & ~35.0h  & ~~~~~~~1.25 \second & 31.01 & 0.947 & 0.081 & -\\
            Plenoxels\cite{yu2021plenoxels} & 5000 & 128k & 11.4m  & 194.50 & 31.71 & 0.958 & 0.049 & - \\
            DVGO\cite{sun2022direct} & 5000 & 30k & 15.0m  &  153.00 & 31.95 & 0.960 & 0.053 & 0.035 \\
            Point-NeRF$_{200k}$\cite{xu2022point} & 4096 & 200k & ~~5.5h & ~~27.74 & ~~~~33.31 \second & ~~~0.962~*  & 0.049& 0.027\\ 
            InstantNGP\cite{muller2022instant}& 10k-85k & 30k & ~~3.9m  &  ~~11.64 & 32.59 & 0.960 & - & - \\
            TensoRF-CP\cite{chen2022tensorf}& 4096 & 30k & 25.2m & ~~~~~~~0.98 \first & 31.56 & 0.949 & 0.076 & 0.041 \\
            TensoRF-VM\cite{chen2022tensorf}& 4096 & 30k &  17.4m & ~~17.95 & 33.14 & ~~~~0.963 \second & 0.047 & ~~~~0.027 \third \\ \hline
            Ours-24& 4096 & 30k & 34.3m & ~~~~~~~7.07 \third & ~~~~33.24 \third & ~~~~0.963 \second & 0.046 \second &  ~~~~0.026 \second   \\
            Ours-48& 4096 & 30k & 35.7m & ~~~13.52 & ~~~~33.55 \first  & ~~~~0.965 \first & 0.044 \first & ~~~~0.025 \first   \\ \hline
            \end{tabular}
        }
        \captionsetup{aboveskip=5pt}
        \captionsetup{belowskip=-0pt}
        \caption {Comparisons of our method with other radiance-based models \cite{martin2021nerf,ibrnet,liu2020neural,xu2022point, chen2022tensorf,muller2022instant} on the Synthetic-NeRF dataset \cite{martin2021nerf}. Ours-24 is the one with 24 components while Ours-48 is the one with 48 components. We report the corresponding rendering quality (PSNR, SSIM, and LPIPS), model capacity (number of parameters), and training time, batch size and steps. Our model achieves the best rendering quality with a compact model size. We report PointNeRF's updated SSIM.}
        \label{tb:nerfsynth} 
    \end{adjustwidth}
\end{table*}
    
\section{Rendering and Reconstruction}
\label{sec:renrecon}
\paragraph{Volume Rendering}
We evaluates each pixel's color with physically-based volume rendering via differentiable ray marching. Following NeRF \cite{mildenhall2020nerf}, we sample $\ShadNum$ shading points at $\{\ShadingX_\iS\;|\;\iS=1,...,\ShadNum\}$ along the ray, and accumulate radiance by density:
    \begin{equation}
    \begin{aligned}
        \Color &=  \sum^\ShadNum_{\iS=1} \Trans_\iS (1-\exp (-\Dens_\iS \Step_\iS)) \Rad_\iS, \\
        \Trans_\iS &= \exp (-\sum_{\iSS=1}^{\iS-1} \Dens_\iSS \Step_\iSS ).
    \label{eq:raymarching}
    \end{aligned}
    \end{equation}
$\Dens_\iS$ and $\Rad_\iS$ are the density and radiance at shading points; $\Step_\iSS$ is the marching distance per step; $\Trans$ denotes transmittance.

\paragraph{Distributing local tensors.} First of all, to better leverage the sparsity of a scene, we first obtain a geometric prior that roughly covers the scene geometry. 
The geometric prior can be in any commonly-used form, e.g., point cloud, occupancy grid, octree, or mesh vertices. Then we can uniformly distribute tensors in the spatial neighborhood of the geometry. For a multi-scale model, each of the scale will be distributed independently. 
For most of our results, we quickly optimize a coarse RGBA volume from the multi-view images and use the optimized occupancy grid as the geometry prior, as done in DVGO~\cite{sun2022direct}, which finishes in seconds. 

To maintain training stability and speed, each tensor $\TensorThreeD$'s position $\TensorX$ and coverage $\TensorSpace$ is fixed once determined. We also initialize the $3(\TensorCompNumber_\Dens + \TensorCompNumber_\Rad)$ vectors ($\TensorV_{\Dens,r}^X,...,\TensorV_{\Rad,r}^X,...$) of each tensor by normal distribution. For each scale $\iScale$, a $P \times \TensorCompNumber_\Rad$ appearance matrix $\AppearanceMC_\iScale$ is shared by all tri-vector tensors of that scale. Specifically, ``$\AppearanceMC \cdot ()$'' in Eqn.\ref{eq:bours} can be efficiently implemented as a fully-connected neural layer. Therefore, $\AppearanceMC$ for each scale and a global appearance MLP $\AppearanceFunc$ will be implemented as neural networks and initialized by default methods \cite{he2015delving}.

\paragraph{Optimization and objectives.}
Given a set of multi-view RGB images with camera poses, the sparse tri-vector radiance field is per-scene optimized to reconstruct the radiance fields, under the supervision of  the ground truth pixel colors. Following the volume rendering equation \ref{eq:raymarching}, the L2 rendering loss can be past back to the global MLP and  aggregated features, then, all the way to the the appearance matrices and the feature vectors of local tensors. 

We apply a rendering loss to supervise the reconstruction and also apply an $L1$ regularization loss on density feature vectors $\TensorV_{\Dens,r}$ to promote geometry sparsity and to avoid overfitting as done in TensoRF \cite{chen2022tensorf}:
        \begin{align}
            \mathcal{L}_{r} = \lVert C - \GTColor \rVert^2_2,
            \label{eq:lr} \\
            \mathcal{L}_{L1} = \cfrac{1}{N} \sum_{r}^{\TensorCompNumber_\Dens}{\lVert \TensorV_{\Dens,r} \rVert},
            \label{eq:l1}
        \end{align}
where $\GTColor$ is the ground truth pixel color, $\lVert \TensorV_{\Dens,r} \rVert$ is the sum of absolute values of elements on density vectors, and $N$ is the total number of elements.
The total loss is:
        \begin{align}
            \mathcal{L} = \mathcal{L}_{r} + \alpha \mathcal{L}_{L1}.
            \label{eq:ltotal}
        \end{align}
We set the weight of the sparsity term $\alpha$ as $1e^{-5}$ by default. 


\newcommand{\DVGORes}{100^3}
\newcommand{\DVGODuration}{30}

\section{Implementation}

To obtain coarse scene geometry, we modify the coarse density estimation introduced in \cite{sun2022direct} and get a $\DVGORes$ occupancy volume in  $\DVGODuration$ seconds. We can skip this step if there exists available geometric data, e.g., the meshes in ScanNet \cite{dai2017scannet}, or point clouds from multiview stereo. According to the experiments, our method is not very sensitive to the initial geometry. Please refer to Appendix.B. for more details. We set the default number of scales to 3. In a scene box of [-1,1], we rasterize the scene geometry (occuppied voxels centers or points) onto 3 grids with different voxel sizes, e.g., $0.4^3,0.2^3,0.1^3$. For each grid, we distribute tri-vector tensors at the center of its occupied voxels. The tensor spatial coverage of these 3 scales is $0.6^3, 0.3^3, 0.15^3$, respectively. For each scale, we query $\QueryNum = 4$ nearby tensors. Following \cite{sun2022direct}, our feature decoding network $\AppearanceFunc$ is a 2-layer MLP with 128 channels. For each scale, its appearance matrix $\AppearanceMC$ is implemented by a single linear layer of 27 channels. 

We implement the framework in PyTorch \cite{imambi2021pytorch} with customized CUDA operations. During optimization, we adopt the coarse to fine strategy in \cite{chen2022tensorf}, linearly up-sample the vectors' dimension ($I,J,K$) from 29 to 121 for scale one, 15 to 61 for scale two, and 7 to 31 for scale three. The up-sampling happens at step 2000, 3000, 4000, 5500, and 7000. We use the Adam optimizer \cite{kingma2014adam} with initial learning rates of 0.02 for vectors and 0.001 for networks. On a single 3090 RTX GPU, we train each model for 30000 steps while each batch contains 4096 rays. Please find more details in the supplemental materials.


    \begin{figure*}
        \begin{adjustwidth}{-0pt}{0pt}
            \begin{center}
                \includegraphics[width=0.95\textwidth]{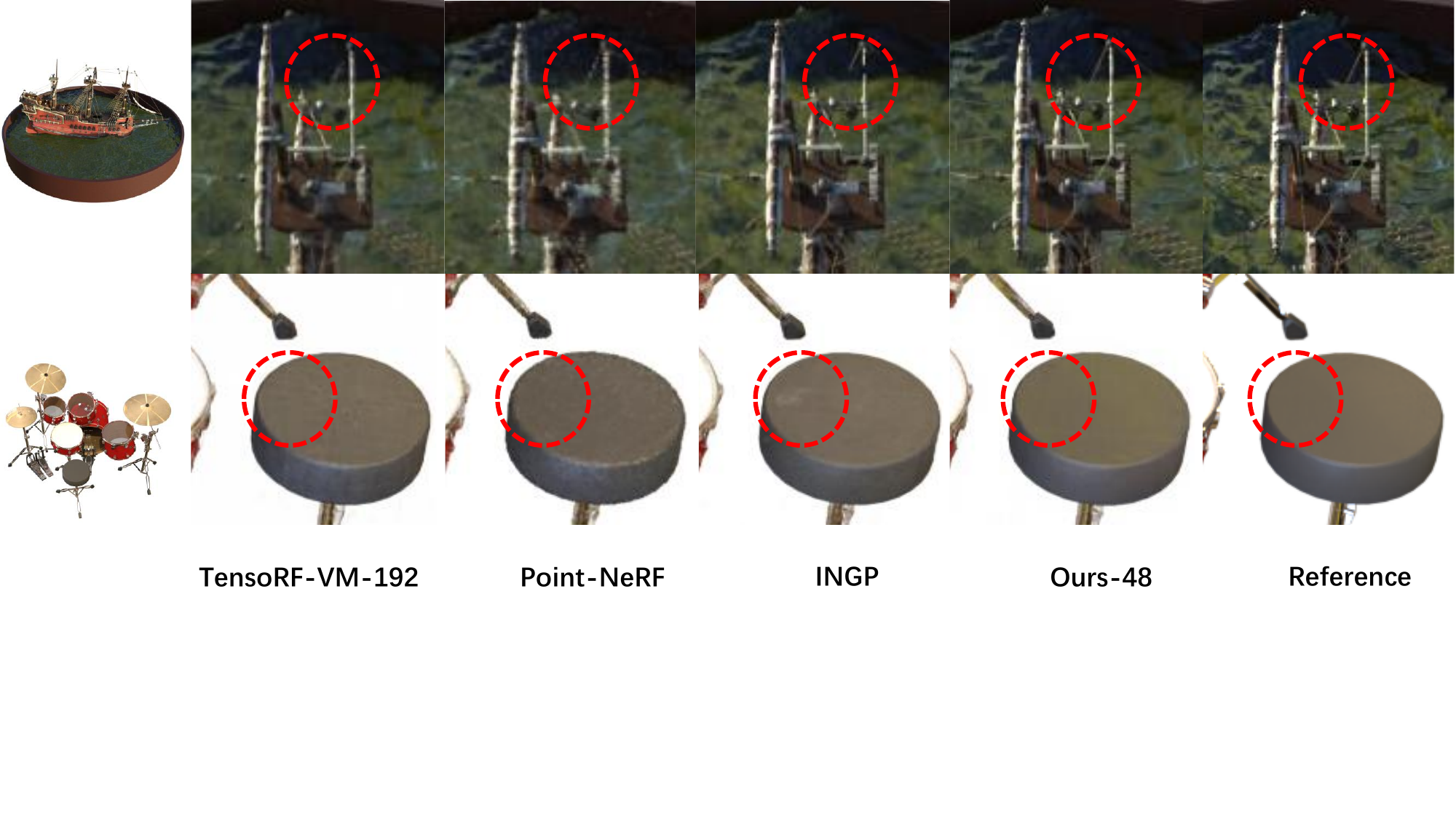}
            \end{center}
        \end{adjustwidth}
         \captionsetup{aboveskip=3pt}
        \captionsetup{belowskip=-5pt}
          \vspace{-65pt}
        \caption{Qualitative comparisons on the NeRF Synthetic dataset \cite{mildenhall2020nerf}.}
        \label{fig:syn_comp}
    \end{figure*}

    \section{Experiments}
    
    \subsection{Evaluation on the NeRF Synthetic Dataset.}
    We first evaluate our method on the Synthetic-NeRF dataset \cite{martin2021nerf} and the quantitative results compared with other methods are reported in Tab.\ref{tb:nerfsynth}, including NeRF \cite{mildenhall2020nerf}, Plenoxels \cite{yu2021plenoxels}, DVGO \cite{sun2022direct}, Point-NeRF \cite{xu2022point}, iNGP \cite{muller2022instant}, and TensoRF \cite{chen2022tensorf}. 
    We report our models of two different model sizes with different numbers of components; both settings are with the same 3 scales of local tensors. 
    
    Our approach achieves the best averaged PSNR, LPIPS$_{Vgg}$ and LPIPS$_{Alex}$ in all the methods, leading to superior visual quality as shown in Fig.~\ref{fig:syn_comp}
    Meanwhile, our high rendering quality is achieved with a compact model size. 
    When compared with local voxel-based representations, such as Plenoxels and DVGO, our approach are significantly better.
    
    On the other hand, global featuring encoding-based methods, like iNGP and TensoRF, are known for their high compactness and can achieve higher rendering quality than local voxel-based methods. 
    Nonetheless, our method can still outperform them. 
    Note that, even our smaller model (Ours-24) leads to better rendering quality than both iNGP and TensoRF that leverage global feature encoding, while our model uses significantly fewer parameters (about 60\% and 40\% of the size of iNGP and TensoRF).
    This clearly demonstrates the high visual quality and high compatness of our model with our sparsely distributed tri-vector tensors.

    In all the baseline methods, Point-NeRF is able to achieve relatively higher rendering quality than others. However, this is enabled by optimizing their model for 300k steps with a long period of 5.5 hours. In contrast, our method achieves higher quality with significantly fewer optimization steps (only 30k) and optimization time (about 36 min).
    As expected, our model is slower to optimize than TensoRF due to the additional costs of multi-tensor aggregation.
    However, though speed is not our focus in this work, our model can still converge quickly and lead to significantly faster reconstruction than MLP-based methods, such as NeRF, as well as Point-NeRF that is point-based.

\begin{figure}[!hbt]
        \centering
        \setlength{\abovedisplayskip}{0pt}%
        \setlength{\abovedisplayshortskip}{\abovedisplayskip}%
        \setlength{\belowdisplayskip}{0pt}%
            \begin{center}{}
                \centering
                \includegraphics[width=1\linewidth]{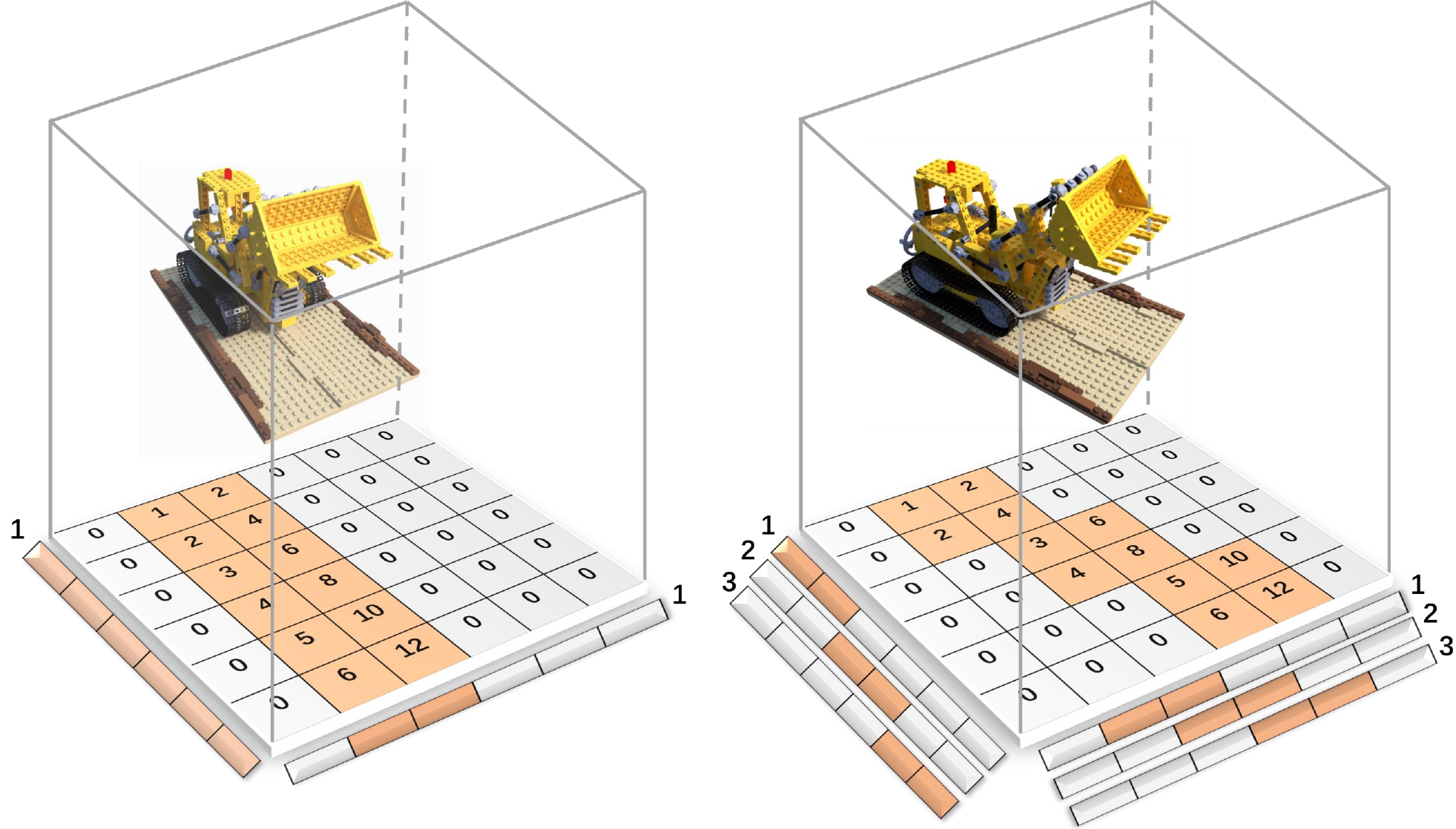} 
            \end{center} 
        \captionsetup{aboveskip=0pt}
        \captionsetup{belowskip=-5pt}
        \caption{A toy example to illustrate the TensoRF-CP with global decomposition in (left) axis-aligned and (right) non-axis-aligned situations. The bottom shows the grid values. In axis-aligned case, only 1 component is needed to represent the scene (vector bases recover grid values by outer product). In non-axis-aligned case, however, 3 components are needed because the rank of matrix changes from 1 to 3 after scene rotation. While our design with local low-rank tensors can alleviate this issue.} 
        \label{fig:toy-rot}
 \end{figure}

    \boldstartspace{Performance w.r.t. rotation.} 
    We observe that tensor factorization-based methods can be sensitive to the orientation of the coordinate frames, since axis-aligned features are used; in essence, this is because the rank of a sparse tensor is sensitive to rotation, as shown in Fig.~\ref{fig:toy-rot}.
    Especially, this can benefit reconstruction on synthetic synthetic scenes where the objects are perfectly aligned with the axes, e.g. the lego scene in the NeRF synthetic data.
    However, we find that our method based on local tensors are more robust against the orientation of the axes than a global TensoRF.
    In particular, we compare our method with TensoRF in Tab.\ref{tb:rot} with different degrees of rotation around the Z axis on two scenes, lego (which is strongly aligned with axes) and chair (which is less aligned and thus less affected ) . 
    As shown in the table, while both methods are affected by the rotation, our method has much smaller drops of PSNRs.

 \begin{table}[hbt]
        \begin{adjustwidth}{0pt}{0pt}  
        \centering
            \small{
                \begin{tabular}{l|ccc}
                \hline 
                chair~/~lego & rot 0$^{\circ}$ & rot 5$^{\circ}$ & rot 45$^{\circ}$ \\ \hline
                TensoRF-CP &  33.60~/~34.50 & 32.90~/~29.79 & 32.50~/~28.57 \\ 
                TensoRF-VM & 35.76~/~36.46 & 34.91~/32.53 & 34.55~/~32.31 \\
                Ours-48&  35.88~/~36.52 & 35.72~/~35.37 & 35.64~/~34.97 \\
                \hline
                \end{tabular}
            }
             \captionsetup{aboveskip=3pt}
            \captionsetup{belowskip=-5pt}
            \caption {Comparisons of our method with TensoRF \cite{ chen2022tensorf} on the chair and lego scenes of Synthetic-NeRF dataset \cite{martin2021nerf} when considering rotation of different angles around $z$-axis.}
            \label{tb:rot} 
        \end{adjustwidth}
    \end{table}

    \begin{figure*}
        \begin{adjustwidth}{-0pt}{0pt}
            \begin{center}
                \includegraphics[width=0.9\textwidth]{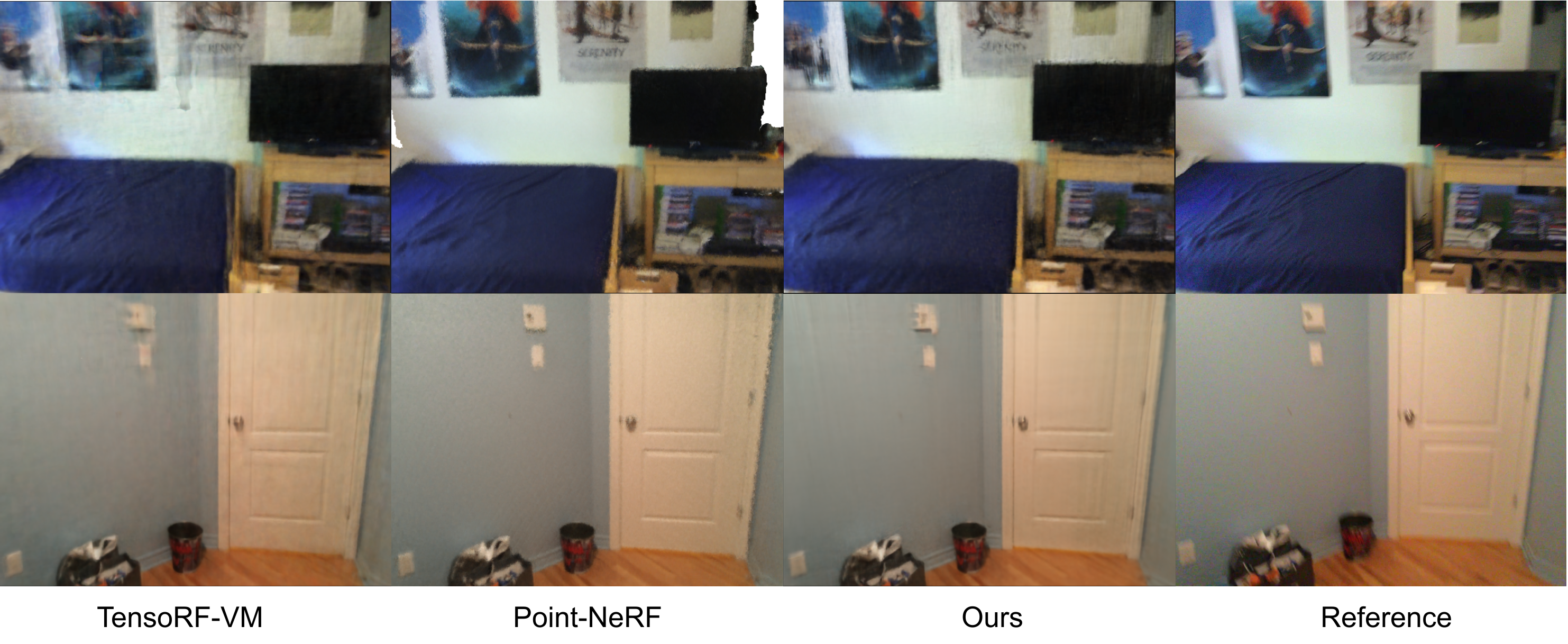}
            \end{center}
        \end{adjustwidth}
        \captionsetup{aboveskip=3pt}
        \captionsetup{belowskip=-10pt}
        \caption{Qualitative comparisons on the ScanNet dataset. } 
        \label{fig:scan_vis}
    \end{figure*}

\subsection{Evaluation on the real datasets.}

    \begin{table}
        \begin{adjustwidth}{0pt}{0pt}  
        \centering
        \setlength\tabcolsep{0.5pt}
            \small{
                \begin{tabular}{lcccc}
                & \multicolumn{4}{c}{\textbf{Average over Scene 101 and Scene 241}}  \\
                \hline 
                \multicolumn{1}{l|}{} & PSNR~$\uparrow$ & SSIM~$\uparrow$ & LPIPS$_{Alex}$~$\downarrow$  & \# Param.(M)~$\downarrow$  \\ \hline
                 \multicolumn{1}{l|}{SRN~\cite{sitzmann2019scene}}  & 18.25 & 0.592 & 0.586 & -\\
                 \multicolumn{1}{l|}{NeRF~\cite{martin2021nerf}}& 22.99 & 0.620 & 0.369 & - \\
                 \multicolumn{1}{l|}{NSVF~\cite{liu2020neural}} & 25.48 & 0.688 & 0.301 & -\\
                 \multicolumn{1}{l|}{Point-NeRF\cite{xu2022point}} &  25.92 & ~~~~~0.891 \first & ~~~~~0.273 \second & 159.01 \\
                 \multicolumn{1}{l|}{TensoRF-CP\cite{chen2022tensorf}} &  ~~~~~27.54 \third & 0.751 & 0.328 & ~~~~~0.97 \first  \\ 
                 \multicolumn{1}{l|}{TensoRF-VM\cite{chen2022tensorf}} & ~~~~~28.61 \second & ~~~~~0.787 \third & ~~~~~0.290 \third & ~~~~~49.92 \third \\ 
                 \multicolumn{1}{l|}{Ours-48} & ~~~~~29.05 \first & ~~~~~0.792 \second & ~~~~~0.243 \first & ~~~~~12.82 \second \\ \hline
                \end{tabular}
            }
             \captionsetup{aboveskip=3pt}
            \captionsetup{belowskip=-5pt}
            \caption {Quantitative comparison on two scenes in the ScanNet dataset \cite{dai2017scannet}. Point-NeRF, TensoRF-CP, TensoRF-VM and Ours-48 are optimized for 100k steps.}
            \label{tb:scannet2} 
        \end{adjustwidth}
    \end{table}
    
\paragraph{The ScanNet dataset.} We evaluate our method on the real dataset, ScanNet \cite{dai2017scannet} with the two scenes selected by NSVF \cite{liu2020neural}, and compare with other methods.
We follow the same experiment setting as done in NSVF \cite{liu2020neural}, using the provided mesh to distribute our tensors, and optimize our model, TensoRF for the same 100k steps for fair comparisons. Please note that Point-NeRF uses all scanned depth images as initial geometry instead of meshes. Therefore, we also obtain the results of Point-NeRF 100k steps from the authors, using the provided mesh for fairness.
We find the Scannet data has many holes in the provided mesh geometry, while methods, such as NSVF and Point-NeRF, require accurate initial geometry; though Point-NeRF can potentially fix them with it point growing technique as shown in their original paper, it is not able to fully address them in 100k step optimization and lead to holes in their final rendering.
Our approach instead does not require very accurate coarse geometry, since our local tensors cover relatively large regions.
We show the quantitative results in Tab.~\ref{tb:scannet2} and qualitative results in Fig.~\ref{fig:scan_vis}.
Note that our method can also perform well on real scenes, achieving the highest performance in terms of PSNR and LPIPS$_{Alex}$, while using the second smallest model size (the smallest one is TensoRF-CP$_{100k}$). 
Our visual quality is also higher than the comparison methods.

\paragraph{The Tanks and Temples dataset.} We also evaluate our method on another real dataset, Tanks and Temples \cite{dai2017scannet} with the 5 object scenes selected by NSVF \cite{liu2020neural}. We using the very coarse geometries estimated by DVGO\cite{sun2022direct} to distribute our tensors. We follow the same experiment setting as done in TensoRF \cite{chen2022tensorf}, optimizing our model for the same 30k steps for fair comparisons. As is shown in Tab.~\ref{tb:scannet2}, our method outperforms other methods in terms of PSNR, SSIM and LPIPS$_{Alex}$, while using the second smallest model size.

\begin{table}[hbt]
        \begin{adjustwidth}{0pt}{0pt}  
        \centering
        \setlength\tabcolsep{3.0pt}
            \small{
                \begin{tabular}{l|cccc}
                \hline 
                 Scale & PSNR~$\uparrow$ & SSIM~$\uparrow$  &  \# Param.(M)~$\downarrow$ & Time~$\downarrow$ \\ \hline
                 Single(0.6)  & 32.22 & 0.957 & 1.75 & 18.22m \\
                 Single(0.3) &  32.73 & 0.961 & 4.15 & 21.31m\\
                 Single(0.15) & 31.96 & 0.952 & 10.20 & 28.55m\\
                 Multi(0.6, 0.3) & 33.11 & 0.963 & 6.20 & 30.12m\\
                 Multi(0.6, 0.3, 0.15) & 33.55 & 0.965 & 13.52 & 35.70m
                 \\ \hline
                \end{tabular}
            }
             \captionsetup{aboveskip=3pt}
            \captionsetup{belowskip=-5pt}
            \caption {Ablation under different scale settings on Synthetic-NeRF dataset. We select 3 scales of tensors with cube sizes of 0.6, 0.3, and 0.15.}
            \label{tb:scale_ablation} 
        \end{adjustwidth}
    \end{table}
\subsection{Ablation study}
    We analyze our model in terms of different scales in Table.\ref{tb:scale_ablation}, while keeping the number of components the same (here we use 48). 
    The scale here is the size of our local tensors of each axis. We considered 3 different scales, i.e., 0.6, 0.3, and 0.15 respectively as single-scale settings and some of their combinations as multi-scale settings. Note that even with a single scale (0.3), the performance of our method can be comparable with some other methods such as iNGP~\cite{muller2022instant} while ours have less than half of the model size. 
    When increasing the number of scales or decreasing the size of local tensors, our model size will also increase. 
    In general, there is a trade-off of our method between scales and computational consumption (time and size). 
    
    Usually, a smaller scale can lead to better performance, though our method with a scale of 0.15 does not strictly follow because we don't have high-quality input geometry to place these local tensors with a very small size. 
    In fact, according to our per-scene breakdown results on the Synthetic-NeRF dataset (please refer to our supplemental material), single-scale(0.075) can achieve higher performance than single-scale(0.3) and single-scale(0.15) on most scenes, except for ship because it has many thin structures that our coarse reconstruction does not cover.

\begin{table}[hbt]
    \begin{adjustwidth}{0pt}{0pt}  
    \centering
    \setlength\tabcolsep{1.5pt}
        \small{
            \begin{tabular}{lcccc}
            \hline 
            \multicolumn{1}{l|}{} & PSNR~$\uparrow$ & SSIM~$\uparrow$ & LPIPS$_{Alex}$~$\downarrow$  & \# Param.(M)~$\downarrow$  \\ \hline
             \multicolumn{1}{l|}{NeRF~\cite{mildenhall2020nerf}}  & 25.78 & 0.864 & 0.198 & - \\
             \multicolumn{1}{l|}{NSVF~\cite{liu2020neural}} & 28.40 & 0.900 & 0.153 & - \\
             \multicolumn{1}{l|}{TensoRF-CP$_{30k}$\cite{chen2022tensorf}} &  27.59 & 0.897 & 0.144 & \textbf{0.97} \\ 
             \multicolumn{1}{l|}{TensoRF-VM$_{30k}$\cite{chen2022tensorf}} & 28.56 & 0.920 & 0.125 & 49.92 \\ 
             \multicolumn{1}{l|}{Ours-48$_{30k}$} &\textbf{28.70} & \textbf{0.922}  & \textbf{0.113} & 14.11\\ \hline
            \end{tabular}
        }
         \captionsetup{aboveskip=3pt}
        \captionsetup{belowskip=-10pt}
        \caption {Quantitative comparison on scenes in the Tanks and Temples dataset \cite{Knapitsch2017} selected in NSVF \cite{liu2020neural}.  TensoRF-CP, TensoRF-VM and Ours-48 are optimized for 30k steps.}
        \label{tb:tt} 
    \end{adjustwidth}
\end{table}

We also compare our method with a variant that uses vector-matrix (VM) decomposition~\cite{chen2022tensorf} in each local tensor instead of CP decomposition. Please refer to Appendix.A. for more details. Also, we can achieve a higher training and inference speed without a significant performance drop, which we refer to Appendix.E.

\section{Conclusion}
In this work, we have presented a novel approach for high-quality neural scene reconstruction and photo-realistic novel view synthesis.
We propose a novel tensor factorization-based scene representation, which leverages CP decomposition to compactly model a 3D scene as a sparse set of multi-scale tri-vector tensors that express local radiance fields.
Our representation leverages both sparsity and spatial local coherence, and leads to accurate and efficient modeling of complex scene geometry and appearance.
We demonstrate that the sparse tri-vector radiance fields can achieve superior rendering quality than previous state-of-the-art neural representations, including TensoRF and iNGP, while using significantly fewer parameters.

{\small
\bibliographystyle{ieee_fullname}
\bibliography{egbib}
}

\newpage
\begin{appendices}
    \twocolumn[{%
        \renewcommand\twocolumn[1][]{#1}%
        \begin{center}
            \centering
            \LARGE \textbf{\appendixname}
            \vspace{30pt}
        \end{center}%
    }]



\maketitle
\ificcvfinal\thispagestyle{empty}\fi


\section{Ablation Studies on Tensor Factorization Strategies}
    \begin{table}[hbt]
        \begin{adjustwidth}{0pt}{0pt}  
        \centering
        \setlength\tabcolsep{3.0pt}
            \small{
                \begin{tabular}{l|c|ccc}
                \hline 
                  & \# Comp & PSNR~$\uparrow$ & SSIM~$\uparrow$  &  \# Param.(M)~$\downarrow$ \\ \hline
                 Multi(0.6, 0.3, 0.15) & 24 & \textbf{33.24} & \textbf{0.963} & \textbf{7.07} \\
                 Single(0.3) & 96 & 33.02 & 0.963 & 9.15 \\
                 VM-Cloud (0.3) & 6 & 32.59 & 0.959 & 11.36 \\
                 VM-Cloud (0.3) & 12 & 32.99 & 0.962 & 21.64\\
                 \hline
                \end{tabular}
            }
             \captionsetup{aboveskip=3pt}
            \captionsetup{belowskip=-1pt}
            \caption {(a) Comparisons on our method pairing with different factorization strategies, e.g., CP decomposition and vector-matrix (VM) decomposition (row 2 vs 3,4). The local tensors' edge lengths are all set as 0.3. (b) We also compare a single-scale model with a multi-scale model (row 1 vs 2). We evaluate these settings on the NeRF Synthetic dataset~\cite{mildenhall2020nerf} and evaluate them with both rendering quality and model capacity (\#Param. denotes the number of parameters).}
            \label{tb:doubleAblation} 
        \end{adjustwidth}
    \end{table}
Other than CP decomposition, TensoRF \cite{chen2022tensorf} also proposes vector-matrix (VM) decomposition, which factorizes a 3D tensor as the summation of vector-matrix bases. Each basis is the outer product of a matrix along a plane, e.g., the XY plane, and a vector along an orthogonal direction, e.g., the Z axis. For comparison, we also explore to replace our tri-vector representation with the vector-matrix representation for each local tensor. Tab.~\ref{tb:doubleAblation} shows that the single-scale tri-vector cloud can outperform the vector-matrix cloud representation with less model capacity. 

It is not a surprise that our tri-vector cloud representation achieves more compactness. It applies more compression by factorizing each component of a 3D tensor, with a space complexity of $O(IJK)$, into three vectors, with a space complexity of $O(I+J+K)$. On the other hand, vector-matrix cloud representation factorizes it into three vectors and three matrices, which have a space complexity of $O(IJ+JK+IK)$. Even if we reduce the number of components, the vector-matrix clouds still require more space than our tri-vector representations.

In terms of quality, since our method exploits the spatial sparsity of natural scenes, we only need to factorize each local space independently instead of the entire scene together. The more compact tri-vector representation can benefit from the appearance coherence in local space and result in better performance. In TensoRF \cite{chen2022tensorf}, since the entire space is factorized all at once, the radiance information is, in general, less coherent across locations and the CP decomposition will lead to a shortage of rank.

\section{Ablation Studies on Multi-scale Models}
In Tab.\ref{tb:doubleAblation}, we also compare our multi-scale tri-vector radiance fields with the single-scale strategy. In our default model, we have three scales, composed of tensors with lengths 0.15, 0.3, and 0.6, respectively. Similar to the findings in iNGP~\cite{muller2022instant}, our multi-scale models provide more smoothness and lead to a better rendering quality than their single-scale counterparts. The multi-scale model with 24 components (row 1) can already outperform the single-scale model (row 2), which has more parameters.

\section{Ablation Studies on the Number of Tensor Components}

We conduct experiments on the NeRF Synthetic dataset~\cite{mildenhall2020nerf} to show the relationship between rendering performance and the number of tensor components. In Tab.\ref{tb:ablation_nerfsynth}, we compare our multi-scale models with 12, 24, 48, and 96 appearance components, respectively. In general, more tensor components will lead to better performance. We also observe that the benefit of adding more components becomes marginal when the number reaches 48. We speculate that it is harder to learn high-frequency details even though the model's capacity can hold high-rank information. Improvement in this aspect can be a promising future direction.


\begin{table*}
    \begin{adjustwidth}{0pt}{0pt}  
    \centering
        \small{
            \begin{tabular}{c|ccccc }
            \hline 
             & PSNR$\uparrow$ & SSIM$\uparrow$& LPIPS$_{Vgg}$ $\downarrow$ & LPIPS$_{Alex}$ $\downarrow$ & \# Param.(M)$\downarrow$\\ \hline
            Ours-12 & 32.94 & 0.961 & 0.049 & 0.028 & \textbf{4.87}\\
            Ours-24 & 33.24 & 0.963 & 0.046 & 0.026 & 7.07\\
            Ours-48 & 33.55 & 0.965 & 0.044 & 0.025 & 13.52 \\
            Ours-96 & \textbf{33.59} & \textbf{0.965} & \textbf{0.043} & \textbf{0.024} & 21.01  \\ \hline
            \end{tabular}
        }
        \captionsetup{aboveskip=5pt}
        \captionsetup{belowskip=-0pt}
        \caption {Ablation study on the number of tensor components. We use the same setting as our default model but only change the number of components in each variant. These variants are evaluated on the NeRF Synthetic dataset \cite{mildenhall2020nerf}.}
        \label{tb:ablation_nerfsynth} 
    \end{adjustwidth}
\end{table*} 

\section{Ablation Studies on Initial Geometry}
\label{initial_geometry}
We emphasize that our superior quality stems from our novel scene representation rather than the initial geometry.
The initial geometry is simply acquired from a low-res RGBA volume reconstruction, which is coarse and only used to roughly prune empty space.

\begin{figure}[htbp]
\centering
 \includegraphics[width=1\linewidth]{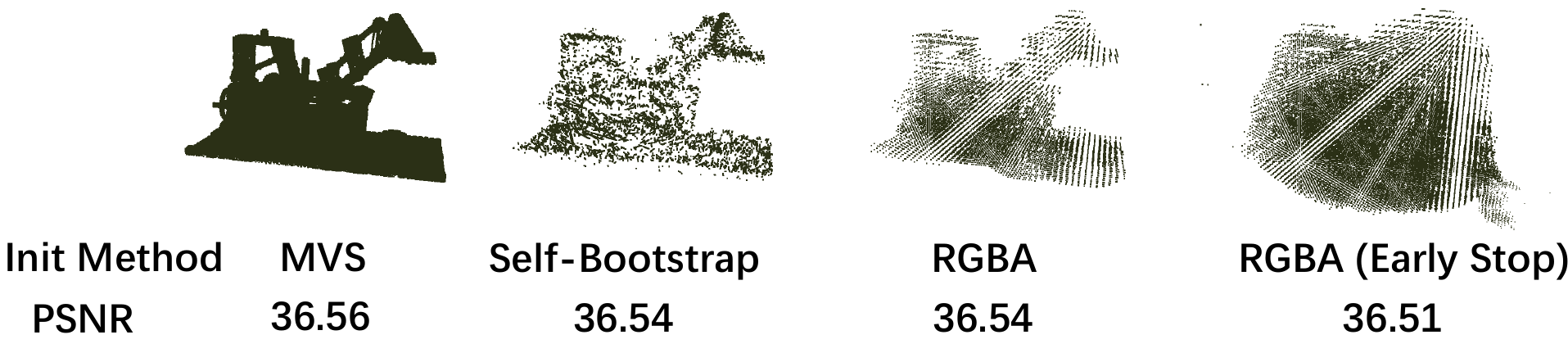}
 \vspace{-6mm}
\caption{Our quality with initial geometry by different methods. }
\label{init_geo}
\vspace{-4mm}
\end{figure}

We show in Fig.~\ref{init_geo} that our approach performs robustly with various choices of these geometry structures and consistently achieves high PSNRs, even with a much worse early-stopped RGBA reconstruction.
This showcases the key to our superior quality is our Strivec model itself.\\
In particular, the self-bootstrap geometry is generated purely from our own model with 8 coarse tri-vectors without existing modules in previous work. 
Moreover, we can also further prune unoccupied tensors during training but we find this leads to similar quality (0.03db difference) and unnecessary extra (+22\%) training time. 
We instead choose to use one single initial geometry to prune empty space in implementation for its simplicity and efficiency.

\section{Speed v.s. Performance}
\label{speed}
Though speed is not our focus, here, if we reduce the number of scales from 3 to 2 and TopK from 4 to 2 (i.e., Multi(0.6, 0.3) with TopK=2), and Strivec becomes faster than CP and close to VM, while still having competitive quality (see Ours-48(fast) in Tab.\ref{tb:speed}). The fewer ranks of our tensor and the less number of TopK to be find for each sample point along a ray lead to less computation, and thus, acceleration. To conclude, Strivec is capable to improve quality, training time and compactness all together with proper hyper-parameters.
\vspace{-6pt}
\begin{table}[hbt]
      \centering
        \setlength\tabcolsep{2pt}
            \small{
                \begin{tabular}{l|cccc}
                \hline 
                 & Train(s)$\downarrow$ & Inference(s/it)$\downarrow$ & \#Params.(M)$\downarrow$ & PSNR$\uparrow$ \\ \hline
                TensoRF-CP & 1914 & 2.01 & 0.98 & 31.56 \\ 
                TensoRF-VM & 915 & 1.60 & 17.95 & 33.14 \\
                Ours-48(fast) & 959  & 1.67 & 6.20 & 33.09\\
                \hline
                \end{tabular}
            }
            \captionsetup{aboveskip = 2pt}
            \captionsetup{belowskip = -15pt}
            \caption {Comparison on NeRF Synthetic dataset~\cite{mildenhall2020nerf}. We compare the average training time (s), inference time (s/it), the number of parameters (M) and PSNR.  }
            \label{tb:speed} 
    \end{table}

\section{Per-scene Breakdown Results of the NeRF Synthetic Dataset}
    We show the per-scene detailed quantitative results of the comparisons on the NeRF Synthetic dataset \cite{mildenhall2020nerf} in Tab. \ref{tb:dt_nerfsynth} 
    and qualitative comparisons in our video. With compact model capacity, our method outperforms state-of-the-art methods \cite{mildenhall2020nerf,muller2022instant,xu2022point,chen2022tensorf} and achieves the best PSNRs, and LPIPSs in most of the scenes. 
    We report two versions of iNGP~\cite{muller2022instant}. Specifically, iNGP-dark$_{100k}$ is reported in the original paper. According to issue~\href{https://github.com/NVlabs/instant-ngp/discussions/745}{\#745} in iNGP's official repo, the method uses a random color background in training and dark background in testing. The number of iterations, 100k, is referenced to its initial code base release. We also refer to the results reported in \cite{factorfields} as iNGP-white$_{30k}$, since the authors use a white background in both training and testing for 30k iterations, which has the same setting as ours and many other compared methods. Please refer to issue \href{https://github.com/NVlabs/instant-ngp/discussions/745}{\#745} and \href{https://github.com/NVlabs/instant-ngp/issues/1266}{\#1266} in iNGP's official repo for more details.
\begin{table}[]
      \centering
            \small{
                \begin{tabular}{l|cc|c}
                \hline 
                 & garden & room & Model Size(avg) \\ \hline
                DVGO & 24.32 & 28.35 & 5.1GB \\ 
                Ours-48 & 24.13 & 28.11 & 12.6MB\\
                \hline
                \end{tabular}
            }
            \captionsetup{aboveskip = 2pt}
            \captionsetup{belowskip = -15pt}
            \caption {Results on the Mip-NeRF 360 dataset.}
            \label{tb:360_tab} 
    \end{table}
    
    \begin{table*}[]
      \captionsetup{aboveskip=5pt}
      \centering
      \begin{tabular}{lcccccccc}
            \hline
            \multicolumn{9}{c}{NeRF Synthetic}                                                                                                                 \\
                       & Chair          & Drums          & Lego           & Mic            & Materials      & Ship           & Hotdog         & Ficus          \\ \hline
            \multicolumn{9}{c}{PSNR$\uparrow$}                                                                                                                           \\ \hline

            NeRF~\cite{mildenhall2020nerf}       & 33.00          & 25.01          & 32.54          & 32.91          & 29.62          & 28.65          & 36.18          & 30.13          \\
            NSVF~\cite{liu2020neural}       & 33.19          & 25.18          & 32.54          & 34.27          & \textbf{32.68} & 27.93          & 37.14 & 31.23          \\
            
            Point-NeRF$_{20k}$~\cite{xu2022point}  & 32.50 & 25.03 & 32.40 & 32.31 &  28.11 & 28.13 & 34.53 & 32.67          \\
            Point-NeRF$_{200k}$~\cite{xu2022point} & 35.40 & 26.06 & 35.04 & 35.95 & 29.61 & 30.97 & 37.30 & 36.13 \\ 
            iNGP-dark$_{100k}$~\cite{muller2022instant} & 35.00 & 26.02 & 36.39 & 36.22 & 29.78 & 31.10 & 37.40 & 33.51 \\
            iNGP-white$_{30k}$~\cite{muller2022instant,chen2023factor} & 35.42 & 24.24 & 34.82 & 35.98 & 28.99 & 30.72 & 37.45 & 32.09 \\ 
            TensoRF-CP~\cite{chen2022tensorf}-384$_{30k}$ & 33.60 & 25.17 & 34.05 & 33.77 & 30.10 & 28.84 & 36.24 &  30.72 \\ 
            TensoRF-VM~\cite{chen2022tensorf}-192$_{30k}$ & 35.76 &  26.01 & 36.46 & 34.61 & 30.12 & 30.77 & 37.41 &  33.99 \\ 
            Ours-12$_{30k}$ & 35.21	& 25.96 & 35.60 & 35.29 & 29.54 & 30.64 & 37.03 & 34.21 \\
            Ours-24$_{30k}$ & 35.60 & 26.16 & 36.05 & 35.81 & 29.79 & 30.89 & 37.24 & 34.37 \\
            Ours-48$_{30k}$ & \textbf{35.88} & \textbf{26.20} & \textbf{36.52} & \textbf{36.65} & 29.90 & \textbf{31.13} & \textbf{37.63} & 34.47 \\ \hline
            \multicolumn{9}{c}{SSIM$\uparrow$}                                                                                                                           \\ \hline
            
            NeRF       & 0.967          & 0.925          & 0.961          & 0.980          & 0.949          & 0.856          & 0.974          & 0.964          \\
            NSVF       & 0.968          & 0.931          & 0.960          & 0.987          & \textbf{0.973}          & 0.854          & 0.980          & 0.973          \\
            
            Point-NeRF$_{20k}$  & 0.981 & 0.944 & 0.980 &  0.986 & 0.959 & 0.916 & 0.983 & 0.986          \\
            Point-NeRF$_{200k}$ & \textbf{0.991} & \textbf{0.954} & \textbf{0.988} & \textbf{0.994} & 0.971 & \textbf{0.942} &  \textbf{0.991} & \textbf{0.993} \\
            
            iNGP-white$_{30k}$ & 0.985 & 0.924 & 0.979 & 0.990 & 0.945 & 0.892 & 0.982 & 0.977 \\
          
             TensoRF-CP-384$_{30k}$ & 0.973 & 0.921 &  0.971 & 0.983 & 0.950 &  0.857 &  0.975 &  0.965  \\
             TensoRF-VM-192$_{30k}$ & 0.985 & 0.937 & 0.983 & 0.988 & 0.952 & 0.895 & 0.982 & 0.982\\
             Ours-12$_{30k}$ & 0.983 & 0.937 & 0.980 & 0.989 & 0.948 & 0.888 & 0.981 & 0.983  \\
             Ours-24$_{30k}$ & 0.984 & 0.940 & 0.982 & 0.990 & 0.952 & 0.893 & 0.982 & 0.984 \\
             Ours-48$_{30k}$ & 0.985 & 0.940 & 0.984 & 0.992 & 0.953 & 0.899 & 0.983 & 0.985 \\\hline
             
            \multicolumn{9}{c}{LPIPS$_{Vgg}\downarrow$}                                                                                                                       \\ \hline
           
            NeRF       & 0.046          & 0.091          & 0.050          & 0.028          & 0.063          & 0.206          & 0.121          & 0.044          \\
            
            Point-NeRF$_{20k}$  & 0.051 & 0.103 & 0.054 & 0.039 & 0.102 & 0.181 & 0.074 & 0.043         \\
            Point-NeRF$_{200k}$ & 0.023 & 0.078 & 0.024 &  0.014 & 0.072 &\textbf{0.124} & 0.037 & 0.022 \\
            iNGP-white$_{30k}$ &  0.022 & 0.092 & 0.025 & 0.017 & 0.069 & 0.137 & 0.037 & 0.026\\

            TensoRF-CP-384$_{30k}$ & 0.044 & 0.114 &  0.038 & 0.035 & 0.068 & 0.196 & 0.052 & 0.058 \\
            TensoRF-VM-192$_{30k}$ & 0.022 & 0.073 & 0.018 & 0.015 & 0.058 & 0.138 & 0.032 & 0.022\\
            Ours-12$_{30k}$ & 0.025 & 0.070 & 0.022 & 0.015 & 0.062 & 0.145 & 0.033 & 0.022 \\
            Ours-24$_{30k}$ &  0.022 & 0.067 & 0.020 & 0.013 & 0.058 & 0.141 & 0.031 & 0.021 \\
            Ours-48$_{30k}$ & \textbf{0.021} & \textbf{0.064} & \textbf{0.017} & \textbf{0.011} & \textbf{0.056} & 0.138 & \textbf{0.029} & \textbf{0.018} \\ \hline
            \multicolumn{9}{c}{LPIPS$_{Alex}\downarrow$}                                                                                                                      \\ \hline
            NSVF       & 0.043          & 0.069          & 0.029          & 0.010          & \textbf{0.021} & 0.162          & 0.025          & 0.017          \\
            Point-NeRF$_{20k}$  & 0.027 & 0.057 & 0.022 & 0.024 & 0.076& 0.127 & 0.044 & 0.022         \\
            Point-NeRF$_{200k}$ & 0.010 & 0.055 & 0.011 & 0.007 & 0.041 & \textbf{0.070} & 0.016 & \textbf{0.009} \\ 

            iNGP-white$_{30k}$ & 0.022 & 0.093 & 0.025 & 0.017 & 0.069 & 0.140 & 0.037 & 0.026 \\
            
            TensoRF-CP-384$_{30k}$ & 0.022 & 0.069 & 0.014 & 0.018 & 0.031 & 0.130 &  0.024 & 0.024 \\
            TensoRF-VM-192$_{30k}$ & 0.010 & 0.051 & 0.007 & 0.009 & 0.026 & 0.085 & 0.013 & 0.012 \\
            Ours-12$_{30k}$ & 0.011	& 0.051 & 0.009 & 0.007 & 0.027 & 0.092 & 0.015 & 0.013  \\
            Ours-24$_{30k}$ & 0.010 & 0.049 & 0.008 & 0.006 & 0.024 & 0.087 & 0.014 & 0.012  \\
            Ours-48$_{30k}$ & \textbf{0.009} & \textbf{0.048} & \textbf{0.007} & \textbf{0.005} & 0.023 & 0.086 & \textbf{0.012} & 0.011 \\ \hline
        \end{tabular}      
        \caption{Detailed breakdown of quantitative metrics on individual scenes in the NeRF Synthetic \cite{mildenhall2020nerf} for our method and baselines. All scores are averaged over the testing images. The subscripts are the number of iterations of the models. NeRF only \cite{mildenhall2020nerf} reports the LPIPS$_{Vgg}$~\cite{zhang2018perceptual} while NSVF only reports LPIPS$_{Alex}$. } 
        \label{tb:dt_nerfsynth}
    \end{table*}

\begin{figure*}
        \begin{adjustwidth}{-0pt}{0pt}
            \begin{center}
                \includegraphics[width=0.9\textwidth]{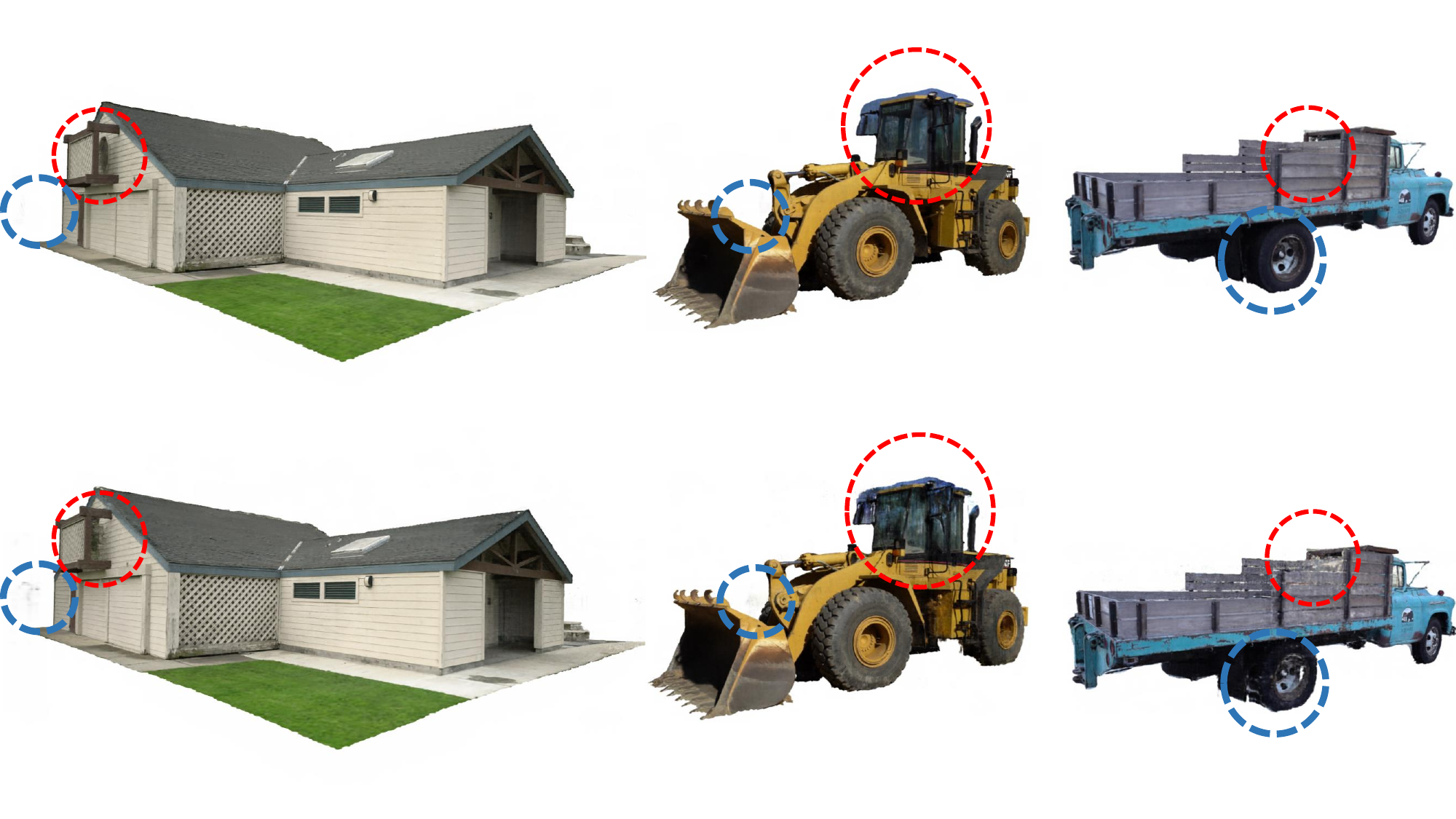}
            \end{center}
        \end{adjustwidth}
        \captionsetup{aboveskip=3pt}
        \captionsetup{belowskip=-10pt}
        \caption{Qualitative comparison on the Tanks and Temples dataset. Top: ours. Bottom: TensoRF-VM.} 
        \label{fig:tanks_comp}
    \end{figure*}

\section{The Tanks and Temples Dataset}
We show the qualitative comparison between our Strivec and TensoRF-VM~\cite{chen2022tensorf} on the Tanks and Temples dataset~\cite{Knapitsch2017} in Fig.\ref{fig:tanks_comp}. Similar to the procedures on the NeRF Synthetic dataset, we build the coarse scene geometry within 30 seconds to place our local tensors. The quantitative results are reported in Tab.\ref{tb:tt}. 
    
    \begin{table*}[hbt!]
      \centering
      \captionsetup{aboveskip=5pt}
        \begin{tabular}{ccccccc}
        \hline
        \multicolumn{7}{c}{Tanks \& Tamples}                                                                                                                \\
        \multicolumn{1}{l}{} & Ignatius             & Truck                & Barn      & Caterpillar                 & Family               & Mean                 \\ \hline
        \multicolumn{1}{l}{} & \multicolumn{1}{l}{} & \multicolumn{1}{l}{} & PSNR~$\uparrow$      & \multicolumn{1}{l}{} & \multicolumn{1}{l}{} & \multicolumn{1}{l}{} \\ \hline
        NV~\cite{lombardi2019neural}                   & 26.54                & 21.71                & 20.82     & 20.71                & 28.72                & 23.70                \\
        NeRF~\cite{mildenhall2020nerf}                 & 25.43                & 25.36                & 24.05     & 23.75                & 30.29                & 25.78                \\
        NSVF~\cite{liu2020neural}                 & 27.91                & 26.92                & 27.16     & 26.44                & 33.58                & 28.40                \\
        TensoRF-CP\cite{chen2022tensorf}    & 27.86                & 26.25                & 26.74     & 24.73                & 32.39                & 27.59 \\
        TensoRF-VM\cite{chen2022tensorf}  & 28.34 & 27.14 & 27.22 & 26.19 & \textbf{33.92} & 28.56 \\
        Ours-48 & \textbf{28.39} &	\textbf{27.32}	& \textbf{28.09} &	\textbf{26.58} &	33.13	& \textbf{28.70}
        \\ \hline
        \multicolumn{1}{l}{} & \multicolumn{1}{l}{} & \multicolumn{1}{l}{} & SSIM~$\uparrow$      & \multicolumn{1}{l}{} & \multicolumn{1}{l}{} &                      \\ \hline
        NV~\cite{lombardi2019neural}                   & 0.992                & 0.793                & 0.721     & 0.819                & 0.916                & 0.848                 \\
        NeRF~\cite{mildenhall2020nerf}                 & 0.920                & 0.860                & 0.750     & 0.860                & 0.932                & 0.864                 \\
        NSVF~\cite{liu2020neural}                 & 0.930                & 0.895                & 0.823     & 0.900                & 0.954                & 0.900                 \\
       TensoRF-CP\cite{chen2022tensorf}    &   0.934 & 0.885 & 0.839 &  0.879 & 0.948 & 0.897 \\
        TensoRF-VM\cite{chen2022tensorf}  &   0.948 & 0.914 &  0.864 & 0.912 & \textbf{0.965} & 0.920\\
        Ours-48 & \textbf{0.948} & \textbf{0.915} & \textbf{0.884} & \textbf{0.917} & 0.957 & \textbf{0.924} \\ \hline
        \multicolumn{1}{l}{} & \multicolumn{1}{l}{} & \multicolumn{1}{l}{} & LPIPS$_{Alex}\downarrow$ & \multicolumn{1}{l}{} & \multicolumn{1}{l}{} &                      \\ \hline
        NV~\cite{lombardi2019neural}                   & 0.117                & 0.312                & 0.479     & 0.280                & 0.111                & 0.260                 \\
        NeRF~\cite{mildenhall2020nerf}                 & 0.111                & 0.192                & 0.395     & 0.196                & 0.098                & 0.198                 \\
        NSVF~\cite{liu2020neural}                 & 0.106                & 0.148                & 0.307     & 0.141                & 0.063                & 0.153                 \\
        TensoRF-CP\cite{chen2022tensorf}  & 0.089 & 0.154 & 0.237 &  0.176 & 0.063 & 0.144\\
        TensoRF-VM\cite{chen2022tensorf}  &  \textbf{0.081} & 0.129 & 0.217 &  0.139 & \textbf{0.057} & 0.125\\
        Ours-48 & 0.083 & \textbf{0.123} & \textbf{0.167} & \textbf{0.125} & 0.065 & \textbf{0.113}\\ \hline
        \multicolumn{1}{l}{} & \multicolumn{1}{l}{} & \multicolumn{1}{l}{} & LPIPS$_{Vgg}\downarrow$  & \multicolumn{1}{l}{} & \multicolumn{1}{l}{} &                      \\ \hline
        TensoRF-CP\cite{chen2022tensorf}  & 0.106 & 0.202 & 0.283 & 0.227 & 0.088 & 0.181\\
        TensoRF-VM\cite{chen2022tensorf}  & \textbf{0.078} & \textbf{0.145} & 0.252 & 0.159 & \textbf{0.064} & 0.140\\
        Ours-48 & 0.083 & 0.150 & \textbf{0.216} & \textbf{0.154} & 0.078 & \textbf{0.136} \\ \hline
        \end{tabular}
        \caption{Quantity comparison on five scenes in the Tanks and Temples dataset \cite{Knapitsch2017} selected in NSVF \cite{liu2020neural}. NV, NeRF, and NSVF have not reported their  LPIPS$_{Vgg}$}
        \label{tb:tt}
    \end{table*}

\section{Mip-NeRF360 Dataset}
\begin{figure*}
        \begin{adjustwidth}{-0pt}{0pt}
            \begin{center}
                \includegraphics[width=0.9\textwidth]{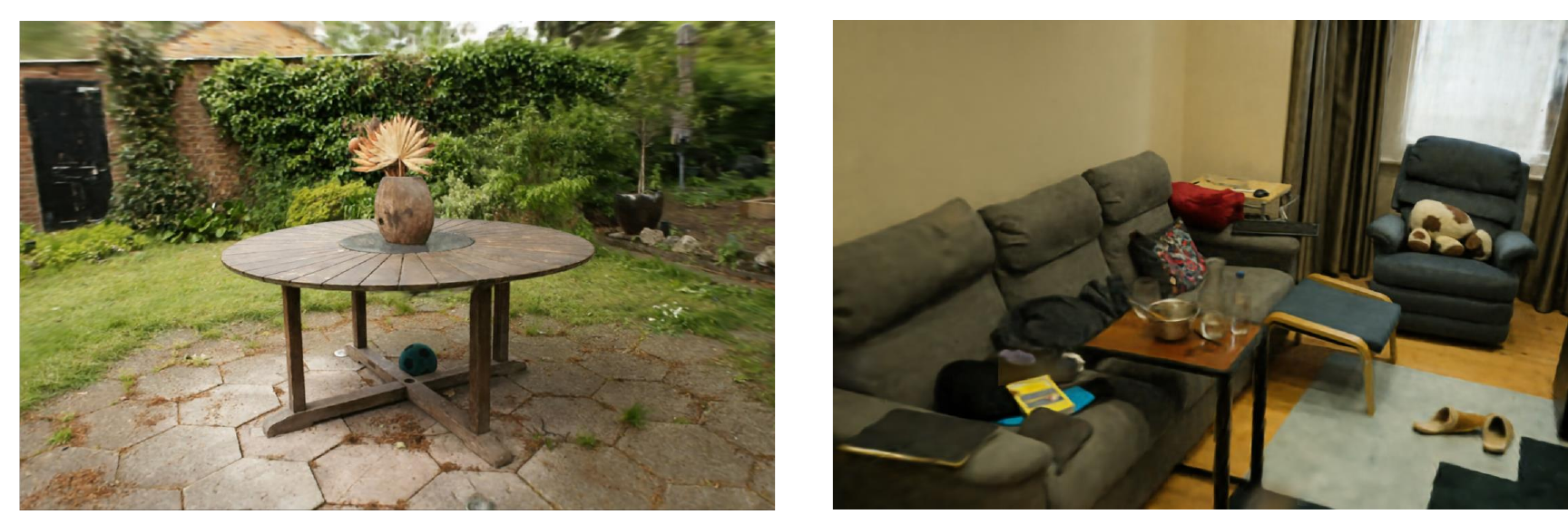}
            \end{center}
        \end{adjustwidth}
        \captionsetup{aboveskip=3pt}
        \captionsetup{belowskip=-10pt}
        \caption{Qualitative results on Mip-NeRF360 dataset. } 
        \label{fig:360_scene}
    \end{figure*}

We evaluate our method on two scenes (one indoor scene and one outdoor scene) of Mip-NeRF360 dataset~\cite{barron2022mip}. Note that we only use the scene warping scheme the same as DVGO~\cite{sun2022direct} and Mip-NeRF360~\cite{barron2022mip} and keeping other components (i.e., positional encoding, point sampling, etc.) the same as TensoRF~\cite{chen2022tensorf}. The qualitative and quantitative results are shown in Fig.~\ref{fig:360_scene} and Tab.~\ref{tb:360_tab} , respectively. Here, we use only two scales in implementation to show our compactness and scalability.  

\begin{figure*}
        \begin{adjustwidth}{-0pt}{0pt}
            \begin{center}
                \includegraphics[width=1\textwidth]{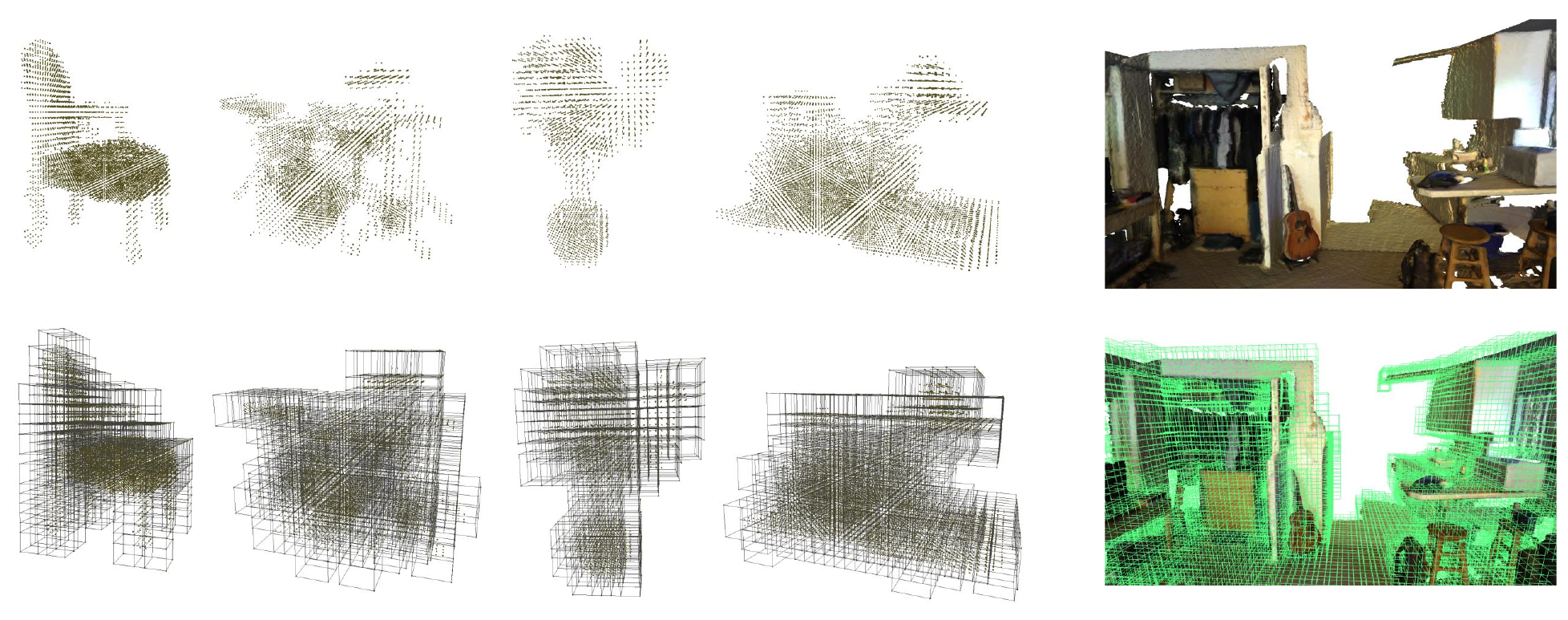}
            \end{center}
        \end{adjustwidth}
        \captionsetup{aboveskip=3pt}
        \captionsetup{belowskip=-10pt}
        \caption{Visualization of local tensors (single scale) on initial geometry. } 
        \label{fig:visual}
    \end{figure*}

\end{appendices}

\end{document}